\documentclass{article}
\usepackage[nonatbib, final]{neurips_2022}
\usepackage{lipsum}
\usepackage[utf8]{inputenc}
\usepackage[T1]{fontenc}    
\usepackage[bookmarks=true, colorlinks=true,linkcolor=magenta,citecolor=green!80!black]{hyperref}
\usepackage[round]{natbib}
\usepackage{wrapfig}
\usepackage{booktabs}  
\usepackage{longtable}
\usepackage{ducksay}
\usepackage{etaremune}
\usepackage{afterpage}
\usepackage{capt-of}
\usepackage[table, x11names]{xcolor}
\usepackage{decorule}
\usepackage{scalerel,xparse}
\usepackage{enumitem} % customization of `list` environments

\usepackage{chngcntr}
\counterwithin{figure}{section} 
\counterwithin{table}{section}

\usepackage{graphicx}
\usepackage{tikz}
\usepackage{tikz-cd}
\usepackage{hf-tikz}
\usepackage{pgfplots} 
\pgfplotsset{compat=1.17} 
\pgfplotsset{
        table/search path={figures/drawings},
    }

\usetikzlibrary{fadings}
\usetikzlibrary{shapes, arrows, fit, backgrounds, arrows.meta}

\usetikzlibrary{matrix}
\usetikzlibrary{shadows.blur}
\usetikzlibrary{patterns, tikzmark}
\usetikzlibrary{decorations.pathreplacing, calc, decorations.markings,}

\usepgfplotslibrary{groupplots}
\usepgfplotslibrary{patchplots}

\definecolor{bg}{gray}{0.97}
\definecolor{olive}{rgb}{0.6, 0.6, 0.2}
\definecolor{sand}{rgb}{0.8666666666666667, 0.8, 0.4666666666666667}
\definecolor{wine}{rgb}{0.5333333333333333, 0.13333333333333333, 0.3333333333333333}
\definecolor{deblue}{RGB}{11,132,147}
\definecolor{ocra}{RGB}{204, 119, 34}

\usepackage{lettrine}
\usepackage{Typocaps}

\LettrineTextFont{\itshape}
\usepackage{amsthm}

\newtheorem{lemma}{Lemma}[section]
\newtheorem{corollary}{Corollary}[section]

\newtheorem{proposition}{Proposition}[section]

\usepackage{minitoc}

\newcommand{\chapref}[1]{\hyperref[#1]{Chapter \ref{#1}}}
\newcommand{\secref}[1]{\hyperref[#1]{Section \ref{#1}}}

\usepackage{marginnote}

\usepackage[many]{tcolorbox}
\usepackage[frozencache,cachedir=minted-cache]{minted}
\tcbuselibrary{listings,breakable,xparse,skins}

\usepackage{newunicodechar}
\newunicodechar{λ}{{$\mathtt\lambda$}}
\newunicodechar{μ}{{$\mathtt\mu$}}

\DeclareTCBListing{mintedbox}{O{}m!O{}}{%
    breakable=true,
    listing engine=listings,
    listing only,
    boxsep=0pt,
    left skip=0pt,
    right skip=0pt,
    left=20pt,
    right=0pt,
    top=3pt,
    bottom=3pt,
    arc=5pt,
    leftrule=0pt,
    rightrule=0pt,
    bottomrule=2pt,
    toprule=2pt,
    colback=bg,
    colframe=brown!70!white,
    enhanced,
    overlay={%
    \begin{tcbclipinterior}
    \fill[brown!20!white] (frame.south west) rectangle ([xshift=16pt]frame.north west);
    \end{tcbclipinterior}},
    #3}

\usepackage{xparse}
\DeclareTColorBox{emphbox}{O{black}O{0cm}}{
    enhanced jigsaw,
    breakable,
    outer arc=0pt,
    arc=0pt,
    colback=white,
    rightrule=0pt,
    toprule=0pt,
    top=0pt,
    right=0pt,
    bottom=0pt,
    bottomrule=0pt,
    colframe=#1,
    enlarge left by=#2,
    width=\linewidth-#2,
}

\DeclareTColorBox{emphbox}{O{black}O{0cm}}{
    empty,
    breakable=true,
    outer arc=0pt,
    arc=0pt,
    rightrule=0pt,
    leftrule=2pt,
    borderline west={2pt}{0pt}{#1},
    toprule=0pt,
    top=0pt,
    right=-3pt,
    bottom=0pt,
    bottomrule=0pt,
    colframe=#1,
    enlarge left by=#2,
    width=\linewidth-#2,
}
\usepackage{amsmath, amssymb, amsfonts, mathtools}
\usepackage{physics}
\usepackage{cancel}
\usepackage{thmtools, thm-restate}
\usepackage{accents}

\DeclareMathOperator{\st}{subject~to}

\DeclareMathOperator*{\topk}{top}

\newcommand{\x}{\times}

\usepackage{pict2e, picture}
\makeatletter
\DeclareRobustCommand{\Arrow}[1][]{%
\check@mathfonts
\if\relax\detokenize{#1}\relax
\settowidth{\dimen@}{$\m@th\rightarrow$}%
\else
\setlength{\dimen@}{#1}%
\fi
\sbox\z@{\usefont{U}{lasy}{m}{n}\symbol{41}}%
\begin{picture}(\dimen@,\ht\z@)
\roundcap
\put(\dimexpr\dimen@-.7\wd\z@,0){\usebox\z@}
\put(0,\fontdimen22\textfont2){\line(1,0){\dimen@}}
\end{picture}%
}
\makeatother

\newcommand{\cD}{\mathcal{D}}

\newcommand{\cN}{\mathcal{N}}

\newcommand{\cT}{\mathcal{T}}

\newcommand{\bC}{\mathbb{C}}

\newcommand{\bE}{\mathbb{E}}

\newcommand{\Id}{\mathbb{I}}

\newcommand{\R}{\mathbb{R}}

\newcommand{\bV}{\mathbb{V}}

\DeclareMathAlphabet{\nummathbb}{U}{BOONDOX-ds}{m}{n}
\newcommand{\0}{\nummathbb{0}}

\makeatletter
\DeclareRobustCommand\widecheck[1]{{\mathpalette\@widecheck{#1}}}
\def\@widecheck#1#2{%
    \setbox\z@\hbox{\m@th$#1#2$}%
    \setbox\tw@\hbox{\m@th$#1%
       \widehat{%
          \vrule\@width\z@\@height\ht\z@
          \vrule\@height\z@\@width\wd\z@}$}%
    \dp\tw@-\ht\z@
    \@tempdima\ht\z@ \advance\@tempdima2\ht\tw@ \divide\@tempdima\thr@@
    \setbox\tw@\hbox{%
       \raise\@tempdima\hbox{\scalebox{1}[-1]{\lower\@tempdima\box
\tw@}}}%
    {\ooalign{\box\tw@ \cr \box\z@}}}
\makeatother

\newcommand{\ourmethod}{$\tt{T1}$}

\everypar{\looseness=-1}

 \allowdisplaybreaks[4]

\usepackage{tabularx}
\usepackage{fontawesome}

\newcolumntype{\CeX}{>{\centering\let\newline\\\arraybackslash}X}
\newcolumntype{\CeP}{>{\raggedleft\arraybackslash}p}

\newcommand{\TextAndSymbol}[2]{%
  \begin{tabularx}{\textwidth}{X >{\raggedleft}>{\raggedright\arraybackslash}X}%
    #1 & #2
  \end{tabularx}%
}
\newcommand{\config}[1]{\TextAndSymbol{%
    \textbf{#1}%
  }{\faCogs}}

\usepackage{thmtools} 
\usepackage{thm-restate}

\usepackage[nameinlink,capitalise]{cleveref}
\crefname{section}{§}{§§}
\Crefname{section}{§}{§§}
\crefname{lemma}{lemma}{lemmas}
\Crefname{lemma}{Lemma}{Lemmas}
\crefname{thm}{theorem}{theorems}
\Crefname{thm}{Theorem}{Theorems}

\title{\textit{Transform Once}\\ Efficient Operator Learning in Frequency Domain}

\author{Michael Poli\thanks{Equal contribution authors. Contact email: \url{poli@stanford.edu}.}\\
	{\normalsize Stanford University}\\
	{\normalsize DiffeqML}\\
    \And
    Stefano Massaroli$^*$\\
	\normalsize Mila\\
	{\normalsize DiffeqML}\\
	\And 
    Federico Berto$^*$\\
	{\normalsize KAIST}\\
	{\normalsize DiffeqML}\\
	\AND\normalsize
    Jinykoo Park\\
    \normalsize	KAIST\\
	\And \normalsize
    Tri Dao\\
    \normalsize	Stanford University\\
	\And \normalsize
	Christopher Ré\\
    \normalsize	Stanford University
	\And \normalsize
	Stefano Ermon\\
    \normalsize	Stanford University \\
    {\normalsize CZ Biohub}\\
}

\begin{document}

%%%%% FRONT MATTER %%%%%
%%%%% FRONT MATTER %%%%%

% Title
\maketitle

\doparttoc
\faketableofcontents
% Abstract
\vspace{-1mm}
\vspace{-5mm}
\begin{abstract}
    % Spectral analysis provides one of the most effective paradigms for information--preserving dimensionality reduction in data: often, a simple description of naturally occurring \textit{signals} can be obtained via few terms of periodic basis functions. Neural operators designed for frequency domain learning -- \textit{frequency domain models} (FDMs) -- are based on complex--valued transforms i.e. \textit{Fourier Transforms} (FT), and layers that perform computation on the spectrum and input data separately. This design introduces considerable computational overhead: for each layer, a forward and inverse FT. Instead, this work introduces a blueprint for frequency domain learning through a single transform: \textit{transform once} (\ourmethod{}). To enable efficient, direct learning in the frequency domain we develop a variance preserving weight initialization scheme and investigate various choices of transforms. Our results noticeably streamline the design process of FDMs, pruning redundant transforms, and leading to speedups of $3$ x to $10$ x that increase with data resolution and model size. We perform extensive experiments on learning to solve partial differential equations, including incompressible Navier--Stokes, turbulent flows around airfoils and high-resolution video of smoke dynamics. \ourmethod{} models improve on the test performance of SOTA FDMs while requiring significantly less computation, with over $20\%$ reduction in predictive error across tasks.

    Spectral analysis provides one of the most effective paradigms for information-preserving dimensionality reduction, as simple descriptions of naturally occurring \textit{signals} are often obtained via few terms of periodic basis functions. In this work, we study deep neural networks designed to harness the structure in frequency domain for efficient learning of long-range correlations in space or time: frequency-domain models (FDMs). Existing FDMs are based on complex-valued transforms i.e. \textit{Fourier Transforms} (FT), and layers that perform computation on the spectrum and input data separately. This design introduces considerable computational overhead: for each layer, a forward and inverse FT. Instead, this work introduces a blueprint for frequency domain learning through a single transform: \textit{transform once} (\ourmethod{}). To enable efficient, direct learning in the frequency domain we derive a variance preserving weight initialization scheme and investigate methods for frequency selection in reduced-order FDMs. Our results noticeably streamline the design process of FDMs, pruning redundant transforms, and leading to speedups of $3$ x to $10$ x that increase with data resolution and model size. We perform extensive experiments on learning the solution operator of spatio-temporal dynamics, including incompressible Navier-Stokes, turbulent flows around airfoils and high-resolution video of smoke. \ourmethod{} models improve on the test performance of FDMs while requiring significantly less computation ($5$ hours instead of $32$ for our large-scale experiment), with over $20\%$ reduction in predictive error across tasks.

\end{abstract}

%%%%% MAIN MATTER %%%%%
%%%%% MAIN MATTER %%%%%
\setlength\abovedisplayshortskip{0pt}
\setlength\belowdisplayshortskip{0pt}
\setlength\abovedisplayskip{1pt}
\setlength\belowdisplayskip{1pt}

% Introduction
\section{Introduction}
\begin{center}
    \small \textit{Nature uses only the longest threads to weave her patterns, so that each small piece of her fabric reveals the organization of the entire tapestry. \citep{feynman1965character}}
\end{center}

Naturally occurring \textit{signals} are often sparse when projected on periodic basis functions \citep{strang1999discrete}.
Central to recently-introduced instances of frequency-domain neural operators \citep{li2020fourier,tran2021factorized}, which we refer to as \textit{frequency-domain models} (FDMs), is the idea of learning to modify specific frequency components of inputs to obtain a desired output in data space. With a hierarchical structure that blends learned transformations on frequency domain coefficients with regular convolutions, FDMs are able to effectively approximate global, long-range dependencies in higher resolution signals without requiring prohibitively deep architectures. 

\newpage

Yet, existing FDMs suffer from several drawbacks:

\begin{itemize}[leftmargin=1cm]
    \item[1.] \textbf{Slow inference:} every layer of an FDM performs a forward and inverse frequency domain transform, introducing a considerable computational overhead. 
    \item[2.] \textbf{Expensive parameter scaling:} each layer of typical FDMs \citep{li2020fourier} performs a long convolution over the inputs by parametrizing it in frequency domain, which scales poorly in the signal resolution.
    \item[3.] \textbf{Incompatibility:} parameter initialization schemes and layers devised to learn directly in data space can be highly suboptimal when introduced without modifications to FDMs.        
\end{itemize}

Despite attempts to improve performance \citep{gupta2021multiwavelet,tran2021factorized}, scaling FDMs to larger data resolutions and model sizes remains fundamentally challenging\footnote{Existing methods to overcome this limitation avoid the frequency domain of inputs, instead introducing an intermediate patch embedding step \citep{guibas2021adaptive,pathak2022fourcastnet}.}.

In this work, we start by posing the question: 

{\centering
\textit{To reap the benefits of learning on frequency domain representations, is it necessary to construct hierarchical deep models that perform forward and inverse frequency transforms at each layer?}
}

We provide the answer in \textit{Transform Once} (\ourmethod{}), a model that builds representations directly in frequency domain, after a \textit{single} forward transform. Each aspect of \ourmethod{} addresses a specific limitation of existing FDMs:
\begin{itemize}[leftmargin=1cm]
    \item[1.] \textbf{Fast:} by performing a single forward transform and optimizing directly on frequency domain coefficients of target data, \ourmethod{} iterations are \textit{at least} $3$x to $10$x faster. When scaling to larger models and higher resolutions, the relative speedups increase as the overhead of each transform grows.
    \item[2.] \textbf{Favourable scaling:} \ourmethod{} employs a single real-valued transform, which we observe to stabilize training and finetuning of deep networks in frequency domain.
    \item[3.] \textbf{Enhanced compatibility:} removing redundant transforms streamlines the design space for \ourmethod{} architectures compared to existing FDMs, allowing direct introduction of optimized layers developed for other applications e.g. UNets \citep{ronneberger2015u}.
    % the unique structure of a neural operator layer requires specialized tricks, with far less opportunities for transfer of current best practices e.g. weight initialization.
\end{itemize}

In \cref{subsec:history} we provide a (short) history on frequency domain approaches in deep learning, followed by background on FDMs \cref{subsec:background}. In \cref{sec:t1rec}, we describe how to train \ourmethod{} directly in frequency domain and motivate the choice of DCT, in \cref{subsec:inverse} we discuss how to choose modes of reduced-order FDMs and in \cref{subsec:init} we introduce a simple variance-preserving weight initialization scheme for all FDMs. Finally, in \cref{sec:experiments} we evaluate \ourmethod{} on a suite of benchmarks related to learning solution operators for a variety of dynamics: incompressible Navier-Stokes, flow around different airfoil geometries, and high-resolution videos of turbulent smoke \citep{eckert2019scalarflow}.

Across tasks, \ourmethod{} is $3\x$ to $10\x$ faster and reduces predictive errors by $20\%$ on average. Training \ourmethod{} models on high resolution videos ($600$ x $1062$) of turbulent dynamics is significantly faster, requiring $5$ hours instead of $32$ hours (FNOs) for the same number of iterations.

% Problem Setting / Background
\section{Related Work and Background}
\subsection{Learning and Frequency Domain: A Short History}\label{subsec:history}
Links between frequency-domain signal processing and neural network architectures have been explored for decades, starting with the original CNN designs \citep{fukushima1982neocognitron}. \cite{mathieu2013fast,rippel2015spectral} proposed replacing convolutions in pixel space with element-wise multiplications in Fourier domain. In the context of learning to solve \textit{partial differential equations} (PDEs), \textit{Fourier Neural Operators} (FNOs) \citep{li2020fourier} popularized the state-of-the-art FDM layer structure: forward transform $\rightarrow$ learned layer $\rightarrow$ inverse transform. Similar architectures had been previously proposed for generic image classification tasks in \citep{pratt2017fcnn,chi2020fast}. Modifications to the basic FNO recipe are provided in \citep{tran2021factorized,guibas2021adaptive,wen2022u}. A frequency domain representation of convolutional weights has also been used for model compression \citep{chen2016compressing}. Fourier features of input \textit{domains} and periodic activation functions play important roles in deep implicit representations \citep{sitzmann2020implicit,dupont2021coin,poli2022self} and general-purpose models \citep{jaegle2021perceiver}.

\subsection{Learning to Solve Differential Equations}
A variety of deep learning approaches have been developed to solve differential equations: neural operators and physics-informed networks \citep{pmlr-v80-long18a,raissi2019physics,lu2019deeponet,karniadakis2021physics}, specialized architectures \citep{wang2020towards,lienen2022learning}, hybrid neural-numerical methods \citep{poli2020hypersolvers,kochkov2021machine,mathiesen2022hyperverlet, berto2022neural}, and FDMs \citep{li2020fourier,tran2021factorized}, the focus of this work.

\subsection{Frequency-Domain Models}\label{subsec:background}
Let $\cD_n$ ($n$-space) to be the set of real-valued discrete signals\footnote{For clarity of exposition, models and algorithms proposed in the paper are introduced without loss of generality for one-dimensional scalar signals (i.e. $\cD_n\equiv\R^n$).} of resolution $N$.
Our objective is to develop efficient neural networks to process discrete signals $x\in\cD_n$,
\[
    x_0, x_1, \dots,x_{N-1},\quad x_n\in\R.
\]

We define a layer of FDMs mapping $x$ to an output signal $\hat y\in\cD_n$ as the structured operator:
\begin{equation}\label{eq:1}
    \tikzstyle{block} = [draw, fill=blue!20!white, rectangle, rounded corners]
    \tikzstyle{sum} = [draw, fill=blue!20, circle, node distance=1cm]
    \begin{tikzpicture}[>=latex', baseline=(current  bounding  box.center)]
        % nodes
        \node at (9,-0.3) {$
        	\begin{aligned}
                    X &= \mathcal T{(x})                  \quad &&\text{Forward Transform} \\ 
            \hat X &= f_\theta(X)               \quad &&\text{Learned Map} \\ 
             \hat x &= \mathcal T^{-1}{(\hat X)}        \quad && \text{Inverse Transform} \\
                \hat y &= \hat x + g(x)   \quad && \text{Residual}
        \end{aligned}
        $};
        \node (x) at (0, 0) {$x$};
        \node (y) at (5.5,0) {$\hat y$};
        \node [block, blur shadow={shadow blur steps=5}] (T) at (1, 0) {$\mathcal{T}$};
        \node (X) at (1.5, .25) {$X$};
        \node [block, blur shadow={shadow blur steps=5}, fill=green!20] (f) at (2, 0) {$f_\theta$};
        \node (X_) at (2.6, .3) {$\hat X$};
        \node [block, blur shadow={shadow blur steps=5}] (T_) at (3.3, 0) {$\mathcal{T}^{-1}$};
        \node (x_) at (4, .26) {$\hat x$};
        \node [sum, blur shadow={shadow blur steps=5}] (sum) at (4.5,0) {};
        \node [block, blur shadow={shadow blur steps=5}, fill=red!20] (g) at (2, -1) {$g$};
        \node [] at (4.5, 0) {$+$};
        % lines
        \draw [->] (x) -- (T);
        \draw [->] (T) -- (f);
        \draw [->] (f) -- (T_);    
        \draw [->] (T_) -- (sum);
        \draw [->] (x) |- (g);    
        \draw [->] (g) -| (sum);
        \draw [->] (sum) -- (y);
    \end{tikzpicture}
\end{equation}
where $\cT$ is an orthogonal (possibly \textit{complex}) linear operator. We denote the $\cT$-transformed $n$-space with $\cD_k$ ($k$-space) so that $\cT:\cD_n\rightarrow\cD_k$. Typically, we assume $\cT$ to be a \textit{Fourier-type} transform\footnote{e.g. \textit{discrete Fourier transform} (DFT), \textit{discrete cosine transform} (DCT), etc.} \citep[Chapter 8]{oppenheim1999discrete} so that the $k$-space corresponds to the \textit{frequency domain} and its elements form the \textit{spectrum} of the input signal $x$. 

The learned parametric map $f_\theta:\cD_k\rightarrow\cD_k$ is the stem of a FDM layer: it maps the $k$-space into itself and is typically chosen to be rank-deficient in the linear case, e.g. $f_\theta(X) = S_m^\top A(\theta) S_m X, ~A(\theta) \in \bC^{m \times m}$ ($m \leq N)$. The matrix $S_m\in\R^{n\times m}$ selects $m$ desired elements of $X$, setting the rest to zero. In the case of frequency domain transforms, this allows \eqref{eq:1} to preserve or modify only specific frequencies of the input signal $x$. 

Residual connections or residual convolutions $g$ \citep{li2020fourier,wen2022u} are optionally added to reintroduce frequency components filtered by $S_m$. A FDM mixes global transformations applied to coefficients of the chosen transform to local transformations $g$ i.e. convolutions with finite kernel sizes. To ensure that such models can approximate generic nonlinear functions, nonlinear activations are introduced after each inverse transform.

\paragraph{Fourier Neural Operators}
Layers of the form \eqref{eq:1} appear in recent FDMs such as \textit{Fourier Neural Operators} (FNOs) \citep{li2020fourier} 
and variants \citep{tran2021factorized, guibas2021adaptive, wen2022u}.

In example, an FNO is recovered from \eqref{eq:1} by letting $\cT$ be a \textit{Discrete Fourier Transform} (DFT)
\begin{equation*}
    % X_k = \frac{1}{\sqrt{N}}\sum_{n=0}^{N-1} x_n e^{-\frac{2\pi j}{N} nk}~~\mapsto ~~\hat X = \theta S_mX~~\mapsto~~\hat x_n = \frac{1}{\sqrt{m}} \sum_{k=0}^{m-1} \hat X_k e^{\frac{2\pi j}{N} kn}  
    \hat x = \cT^{-1}\circ f_\theta\circ \cT (x) = W^* S^\top_m A(\theta) S_m W x
\end{equation*}
where $W\in\bC^{N\x N}$ is the standard $N$-dimensional DFT matrix and $W^*$ its conjugate transpose. The \textit{Discrete Fourier Transforms} (DFTs) is a natural choice of $\cT$ as it can be computed in $O(N\log N)$ via \textit{Fast Fourier Transform} (FFT) algorithms \citep[Chapter 9.2]{oppenheim1999discrete}.

We identify two major limitations of FDMs in the form \eqref{eq:1}; each layer performs $\cT$ and $\cT^{-1}$ and DFTs are complex-valued, resulting in overheads and a restriction of the design space for $f_\theta(X)$.

With \ourmethod{}, we aim to develop an FDM that does not require more than a single $\cT$, while preserving or improving on predictive accuracy. Ideally, the transform in \ourmethod{} should be (1) real-valued, to avoid restrictions in the design space of the architecture and thus retain compatibility with existing pretrained models, (2) universal, to allow the representation of target signals, and (3) approximately sparse or structured, to allow dimensionality reduction.  
\begin{tcolorbox}[enhanced, drop fuzzy shadow, frame hidden]
    \[
    \begin{tikzpicture}[baseline=(current  bounding  box.center)]
        \def\ngon{5}
        \node[draw=none, regular polygon,regular polygon sides=\ngon,minimum size=2.5cm] (p) {};
        \node[] (p1) at (p.corner 1) {$\hat y$};
        \node[] (p2) at (p.corner 2) {$x$};
        \node[] (p3) at (p.corner 3) {$X$};
        \node[] (p4) at (p.corner 4) {$\hat X$};
        \node[] (p5) at (p.corner 5) {$\hat x$};
        \path[-{Straight Barb[right]}] ([xshift=-1.4pt] p2.south) edge node [left] {$\cT$} ([xshift=-1.4pt] p3.north)
        ([xshift=1.4pt] p3.north) edge node [right] {$\cT^{-1}$} ([xshift=1.4pt] p2.south);
        \path[-stealth]
        (p3) edge node [below] {$f_\theta$} (p4)
        (p3.east) edge (p4.west);
        \path[-{Straight Barb[right]}] ([xshift=-1.4pt] p5.south) edge node [left] {$\cT$} ([xshift=-1.4pt] p4.north)
        ([xshift=1.4pt] p4.north) edge node [right] {$\cT^{-1}$} ([xshift=1.4pt] p5.south);
        \draw[-stealth]
        (p2) edge node [above left] {$g$} (p1)
        (p5) edge (p1);
    \end{tikzpicture}
    \qquad\qquad
    \begin{tikzpicture}[baseline=(current  bounding  box.center)]
        \matrix (m) [matrix of math nodes,row sep=3em,column sep=4em,minimum width=2em]
        {
         x & \hat x \\
         X & \hat X \\};
        \path[-stealth]
        (m-2-1.east|-m-2-2) edge node [below] {$S_m^\top A(\theta) S_m$} (m-2-2);
        %(m-1-1) edge node [left] {$W$} (m-2-1)
        %(m-2-2) edge node [right] {$W^*$} (m-1-2);
        %
        \path[-{Straight Barb[right]}] ([xshift=-1.4pt] m-1-1.south) edge node [left] {$W$} ([xshift=-1.4pt] m-2-1.north)
        ([xshift=1.4pt] m-2-1.north) edge node [right] {$W^*$} ([xshift=1.4pt] m-1-1.south)
        ([xshift=-1.4pt] m-1-2.south) edge node [left] {$W$} ([xshift=-1.4pt] m-2-2.north)
        ([xshift=1.4pt] m-2-2.north) edge node [right] {$W^*$} ([xshift=1.4pt] m-1-2.south);
    \end{tikzpicture}
    \]
    Commutative diagrams for FDM layers \eqref{eq:1} and linear FNOs (frequency domain part).
\end{tcolorbox}
%

% Our Method
%
\section{Transform Once: The \ourmethod{} Recipe}\label{sec:t1rec}

With \ourmethod{}, we introduce major modifications to the way FDMs are designed and optimized. In particular, \ourmethod{} is defined, inferred and trained directly in the frequency domain with only a \textbf{single} direct transform required to process data. Hence follows the name: \textit{transform once} (\ourmethod{}).

\paragraph{Direct learning in the frequency domain}

Consider two signals $x\in\cD_n$, $y\in\cD_n$ and suppose there exists a function $\varphi:\cD_n\rightarrow\cD_n$ mapping $x$ to $y$, i.e.
\[
    \quad y = \varphi(x).
\]
Then, there must also exist another function $\psi:\cD_k\rightarrow\cD_k$ that relates the spectra of the two signals, i.e. $Y = \psi(X)$
being $X = \cT(x)$ and $Y = \cT(y)$. In particular,
\[
    \varphi(x) = \cT^{-1} \circ \psi \circ \cT(x)~~\Leftrightarrow~~\cT\circ\varphi(x) = \psi \circ \cT(x)
\]
It follows that, from a learning perspective, we can aim to approximate $\psi$ directly in the $k$-space rather than $\varphi$ in the $n$-space. To do so, we define a learnable parametric function $f_\theta:\cD_k\rightarrow\cD_k$ and train it to minimize the approximation error $J_\theta$ of the output signal spectrum $Y$ in the $k$-space. Given a distribution $p(x)$ of input signals, \ourmethod{} is characterized by the  following nonlinear program
\begin{equation}\label{eq:2}
    \tikzstyle{block} = [draw, fill=blue!20!white, rectangle, rounded corners]
    \tikzstyle{sum} = [draw, fill=blue!20, circle, node distance=1cm]
    \tikzstyle{container} = [draw, rectangle, inner sep=0.15cm, fill=gray!20,minimum height=1cm, dotted, rounded corners, thick]
    \begin{tikzpicture}[>=latex', baseline=(current  bounding  box.center)]
        % nodes
        \node at (-3.5,-0.5) {$
        	\begin{aligned}
        \min_{\theta}~~& \bE_{x,y}\left[\| \cT(y)- \hat Y \|\right]\\
        \st~~& \hat Y = f_{\theta} \circ \cT(x)\\
        ~~& x \sim p(x) \\
        ~~& y = \varphi(x) 
    \end{aligned}
        $};
        \node (x) at (0, 0) {$x$};
        \node (y) at (0, -1) {$y$};
        \node (X) at (1.75, 0) {$X$};
        \node (Y) at (1.75, -1) {$Y$};
        \node (Y_) at (3.35, .3) {$\hat Y$};
        %\node (loss) at (5.5,0) {Loss};
        \node [block, blur shadow={shadow blur steps=5}] (T) at (.8, 0) {$\mathcal{T}$};
        \node [block, blur shadow={shadow blur steps=5}] (T_) at (.8, -1) {$\mathcal{T}$};
        \node [block, blur shadow={shadow blur steps=5}, fill=green!20] (f) at (2.75, 0) {$f_\theta$};
        \node [block, blur shadow={shadow blur steps=5}, fill=yellow!80!black] (J) at (4, 0) {$J_\theta$};
        \begin{scope}[on background layer]
        \node [container,fit=(x) (y) (T) (T_)] (preproc) {};
        \node [container,fit=(X) (Y), fill=blue!20!gray!20] (data) {};
        \end{scope}
        % lines
        \draw [->] (x) -- (T);
        \draw [->] (y) -- (T_);
        %\draw [->] (T) -- (f);
        \draw [-] (T) -- (X);
        \draw [->] (X) -- (f);
        \draw [->] (f) -- (J);
        %\draw [->] (J) -- (loss);
        %\draw [->] (T_) -| (J);
        \draw [-] (T_) -- (Y);
        \draw [->] (Y) -| (J);
        %
        %\node[] at (.85, .85) {offline pre-processing};
    \end{tikzpicture}
\end{equation}
If $\cT$ is a DFT, the above turns out to be a close approximation (or equivalent, depending on the function class of $f_\theta$) to the minimization of $\|y - \hat{y}\|$ in $n$-space by the \textit{Parseval-Plancherel identity}.

\begin{restatable}[Parseval-Plancherel Identity {\cite[pp. 223]{stein2011fourier}} ]{thm}{parseval} \label{prop:parseval}
    Let $\cT$ be the normalized DFT. Given a signal $v\in\cD_n$ and its transform $V = \cT(v)$, it holds $\|v\| = \|V\|$.
\end{restatable}
This result also applies to any other norm-preserving transform $\cT$, e.g.  a normalized type-\textsc{II} DCT \citep[pp. 679]{oppenheim1999discrete}. For the linear transforms considered in this work, $\cT(x) = Wx,~W\in\bC^{N\x N}$, condition for Th. \ref{prop:parseval} to hold is $W$ to be orthonormal, i.e. $W^*W = \Id$.

Note that \ourmethod{} retains, in principle, the same universal approximation properties of FNOs \citep{kovachki2021universal} as $f_\theta$ is allowed to operate on the entirety of the input spectrum. Given enough capacity, $f_\theta$ can arbitrarily approximate $\psi$, implicitly reconstructing $\varphi$ via $\cT^{-1}\circ f_\theta\circ\cT$.

\paragraph{Speedup measurements}
We provide a concrete example of the effect of pruning redundant transforms on computational costs. We measure wall-clock inference time speedups of depth $d$ \ourmethod{}
\[
\text{\ourmethod{}}(x):= f_{d} \circ \cdots\circ f_2 \circ f_1 \circ \cT(x)
\]
over an equivalent depth $d$ FNO with layers \eqref{eq:1}. The only difference concerns the application of transforms between layers. 

\begin{wrapfigure}[17]{r}{0.45\textwidth}\label{clean_speedups}
    \centering
    \vspace{-3mm}
    \includegraphics[width=0.9\linewidth]{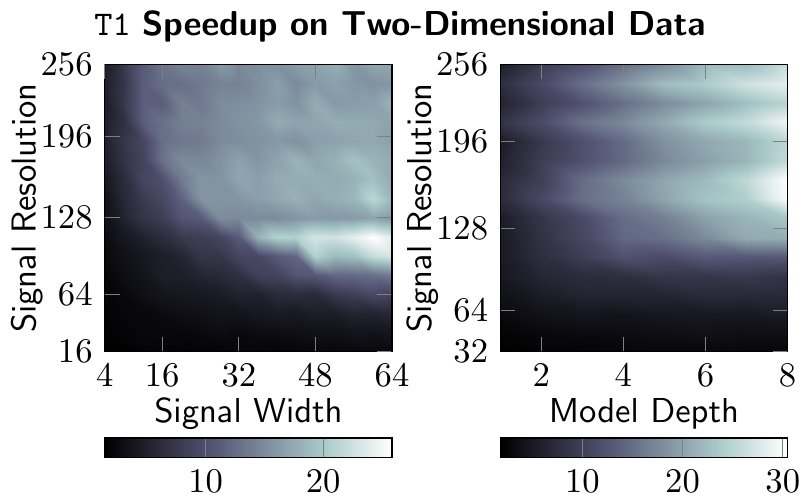}
    \vspace{-5mm}
    \caption{\footnotesize Speedup in a forward pass of \ourmethod{} over FNOs sharing the same transform $\cT$ (DFT) on two-dimensional signals of increasing resolution. The speedup for a given configuration (point on the plane) is shown as background color gradient. The improvement grows with signal width, resolution and model depth. }
    \label{fig:speedup-contour-2d}
\end{wrapfigure}

 \cref{fig:speedup-contour-2d} provides the speedups on two-dimensional signals: on the left, we fix model depth $d=6$ and investigate the scaling in signal width (i.e. number of channels) and signal resolution. On the right, we fix signal width to be $32$ and visualize the interaction of model depth and signal resolution. For common experimental settings e.g. resolutions of $64$ or $128$, $6$ layers and width $32$, \ourmethod{} is at least $10$ x faster than other FDMs. It will later be shown (\cref{sec:experiments}) that \ourmethod{} also preserves or improves on predictive accuracy of other FDMs across tasks.

When \ourmethod{} is not preceded by online preprocessing steps for inputs $x$, such as other neural networks or randomized data augmentations, the transform on $\cT(x)$ can be done once  on the dataset, amortizing the cost over training epochs, and increasing the speed of \ourmethod{} further.

\paragraph{Choosing the right transform}\label{par:choosing}

The transform $\cT$ in \ourmethod{} is chosen to be in the class of \textit{Discrete Cosine Transforms} (DCTs) \citep{ahmed1974discrete,strang1999discrete}, in particular the normalized DCT-II, which can also be computed $\mathcal{O}(N \log N)$ via FFTs \citep{makhoul1980fast}. DCTs provide effective representations of smooth continuous signals \citep{trefethen2019approximation} and are the backbone of modern lossy compression methods for digital images.

Although other transforms are available, we empirically observe DCT-based {\tt T1} to perform best in our experiments. This phenomenon can be explained by the sparsity and energy distribution properties of the transformed spaces, an intrinsic property of the specific dataset and chosen transform. This is in line with results of classic signal processing and compression literature. Particularly, DCT features are known to have a higher energy compaction than their DFT counterparts in a variety of domains, from natural images \citep{yaroslavsky2014fast} to audio signals \citep{soon1998noisy}. Energy compaction is often the decisive factor in choosing a transform for downstream tasks.  

Letting $\mathcal{T}$ be a real-valued transform in {\tt T1} architectures preserves compatibility between $f_{\theta}$ and existing architectures e.g., models pre-trained on natural image datasets.

\subsection{Reduced-Order \ourmethod{} Model and Irreducible Loss Bound}\label{subsec:inverse}

We seek to leverage structure induced in $\cD_k$ by $\cT$. To this end we allow \ourmethod{}, similarly to \eqref{eq:1}, to modify specific elements of $X$ and consequently trasform only certain frequency components of $x$ (and $y$). 

The reduced-order \ourmethod{} model is designed to operate only on $m<N$ elements (selected by $S_m\in\R^{N\times m}$) of the input $k$-space, i.e. on a \textit{reduced} $k$-space $\cD_m\equiv\R^m$ of lower dimension.
Thus, we can employ a smaller neural network $\gamma_\theta:\cD_m\rightarrow\cD_m$ for mapping $S_m X$ to the corresponding $m$ elements $S_m Y$ of the output $k$-space. Thus, training involves a truncated objective that compares predictions with elements in the output signal spectrum also selected by $S_m$:
\begin{equation}\label{eq:3}
    \tikzstyle{block} = [draw, fill=blue!20!white, rectangle, rounded corners,  blur shadow={shadow blur steps=3}, thick]
    \tikzstyle{el} = [draw, fill=gray!20, circle, blur shadow={shadow blur steps=3}, minimum size=.25cm, inner sep=0pt, outer sep=0pt, thick]
    \tikzstyle{container} = [semithick, draw, rectangle, inner sep=0.1cm, rounded corners, fill=blue!10!gray!10]
    \begin{tikzpicture}[>=latex', baseline=(current  bounding  box.center)]
        \node[el] at (0,0) (x0) {};
        \node[el] at (0,-.5) (x1) {};
        \node[el] at (0,-1) (x2) {};
        \node[el] at (0,-1.5) (x3) {};
        \node[el] at (0,-2) (x4) {};
        \node[el] at (0,-2.5) (x5) {};
        \node[el] at (0,-3) (x6) {};
        \node[el] at (.75, 0) (X0) {};
        \node[el, fill=blue!50] at (.75,-.5) (X1) {};
        \node[el, fill=blue!50] at (.75,-1) (X2) {};
        \node[el] at (.75,-1.5) (X3) {};
        \node[el] at (.75,-2) (X4) {};
        \node[el, fill=blue!50] at (.75,-2.5) (X5) {};
        \node[el] at (.75,-3) (X6) {};
        \node[el, fill=blue!50] at (1.5,0) (X0s) {};
        \node[el, fill=blue!50] at (1.5,-.5) (X1s) {};
        \node[el, fill=blue!50] at (1.5,-1) (X2s) {};
        \node[block] at (2.25,-.5) (gamma) {$\gamma_\theta$};
        \node[el, fill=red!50] at (3,0) (Yh0) {};
        \node[el, fill=red!50] at (3,-.5) (Yh1) {};
        \node[el, fill=red!50] at (3,-1) (Yh2) {};
        \node[] at (0,.5) {$x$};
        \node[] at (.75,.5) {$X$};
        \node[] at (1.5,.5) {$S_mX$};
        \node[] at (3,.5) {$\hat Y$};
        \path[-] (x0) edge (X0)
                 (x1) edge (X1)
                 (x2) edge (X2)
                 (x3) edge (X3)
                 (x4) edge (X4)
                 (x5) edge (X5)
                 (x6) edge (X6);
        \path[-] %(X1) edge (X0s)
                 ([xshift=1mm] X1.east) edge ([xshift=-1mm] X1s.west);
                 %(X5) edge (X2s);
        \path[-] (X0s) edge (gamma.west)
                 (X1s) edge (gamma.west)
                 (X2s) edge (gamma.west);
        \path[-] (gamma.east) edge (Yh0)
                 (gamma.east) edge (Yh1)
                 (gamma.east) edge (Yh2);
        \def\x{5};
        \def\y{-3};
        \node[el] at (\x,\y) (y0) {};
        \node[el] at (\x-.5,\y) (y1) {};
        \node[el] at (\x-1,\y) (y2) {};
        \node[el] at (\x-1.5,\y) (y3) {};
        \node[el] at (\x-2,\y) (y4) {};
        \node[el] at (\x-2.5,\y) (y5) {};
        \node[el] at (\x-3,\y) (y6) {};
        \node[el] at (\x + 0,\y+.75) (Y0) {};
        \node[el, fill=green!50] at (\x -.5,\y+.75) (Y1) {};
        \node[el, fill=green!50] at (\x -1,\y+.75) (Y2) {};
        \node[el] at (\x -1.5,\y+.75) (Y3) {};
        \node[el] at (\x -2,\y+.75) (Y4) {};
        \node[el, fill=green!50] at (\x -2.5,\y+.75) (Y5) {};
        \node[el] at (\x -3,\y+.75) (Y6) {};
        \node[el, fill=green!50] at (\x ,\y+1.5) (Y0s) {};
        \node[el, fill=green!50] at (\x -.5,\y+1.5) (Y1s) {};
        \node[el, fill=green!50] at (\x -1,\y+1.5) (Y2s) {};
        \node[] at (\x+.5, \y) {$y$};
        \node[] at (\x+.5, \y+.75) {$Y$};
        \node[] at (\x+.75, \y+1.5) {$S_mY$};
        \path[-] (y0) edge (Y0)
                 (y1) edge (Y1)
                 (y2) edge (Y2)
                 (y3) edge (Y3)
                 (y4) edge (Y4)
                 (y5) edge (Y5)
                 (y6) edge (Y6);
        \path[-] ([yshift=1mm] Y1.north) edge ([yshift=-1mm] Y1s.south);
        
        \node[el, fill=yellow!70!black] (J) at (\x-.5,-.5) {};
        \node[] at (\x,-.5) {$J_\theta$};
        
        \path[->] ([yshift=1mm] Y1s.north) edge ( J.south);
        \path[->] ([xshift=1mm] Yh1.east) edge ( J.west);
        
        %%%%%%%%%%%%%%%
         %
        \begin{scope}[on background layer]
            \node [container,fit=(X0s) (X1s) (X2s)] {};
            \node [container,fit=(Yh0) (Yh1) (Yh2)] {};
            \node [container,fit=(X0) (X1) (X2) (X3) (X4) (X5) (X6)] {};
            \node [container,fit=(x0) (x1) (x2) (x3) (x4) (x5) (x6)] {};
            \node [container,fit=(Y0s) (Y1s) (Y2s)] {};
            \node [container,fit=(Y0) (Y1) (Y2) (Y3) (Y4) (Y5) (Y6)] {};
            \node [container,fit=(y0) (y1) (y2) (y3) (y4) (y5) (y6)] {};
        \end{scope}
        %
        %%%%%%%%%%%%%%%
        \node at (-3.5,-1)
        {$
        \begin{aligned}
            \min_{\theta}~~& \bE_{x,y}\left[\| S_m \circ \cT(y) - \hat Y \|\right]\\
            \st~~& \hat Y = \gamma_{\theta} \circ S_m \circ \cT(x)\\
            ~~& x \sim p(x) \\
            ~~& y = \varphi(x)
        \end{aligned}
        $};
    \end{tikzpicture}
\end{equation}
%

\begin{figure}[b]
    \centering
    \includegraphics[width=\linewidth]{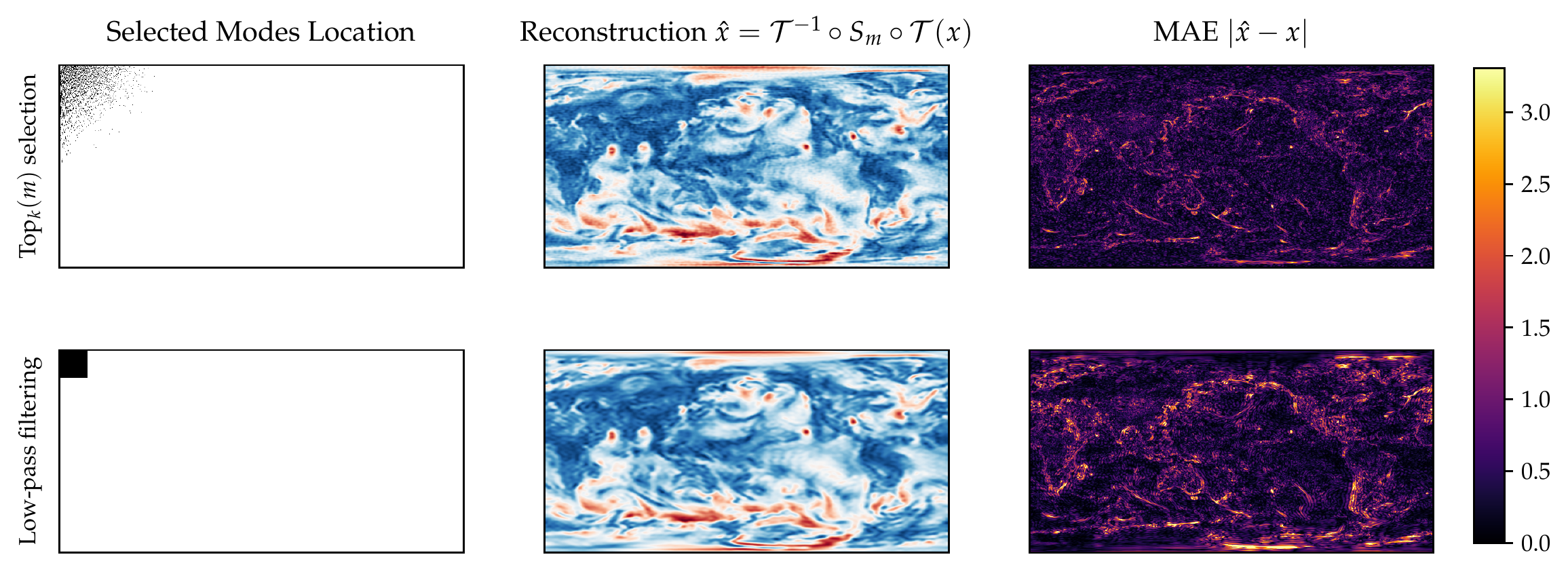}
    \vspace{-5.5mm}
    \caption{Reconstructions after low-pass filtering (first $m$ modes) \textbf{[Bottom]} or top-$m$ selection \textbf{[Top]} of ERA5 \citep{hersbach2020era5} climate data. The non-monotonic structure of the spectrum implies more accurate reconstructions can be obtained with top-$m$ selection.}
    \vspace{-5mm}
    \label{fig:era5}
\end{figure}

% with $S_m:\bR^N\rightarrow\bC^m$ and $\gamma_{\theta}:\bC^{m}\rightarrow\bC^m$, $\gamma_\theta = (\gamma_{\theta,0},\cdots,\gamma_{\theta,m-1})$. 
\clearpage
\paragraph{How to choose modes in reduced-order FDMs} We now detail some intuitions and heuristic to choose which modes $k_0, \dots, k_{m-1}$ should be kept to maximize the information content in the truncated spectrum. For this reason, we evaluate the \textit{irreducible} loss arising from discarding some $N-m$ modes. We recall that the (reduced) $k$-space training objective $J_\theta(X, Y)$ reads as
\[
    J_\theta(X, Y) = \|S_m Y - \hat Y\| = \sum_{l=1}^{m}\left|Y_{k_l} - \gamma_{\theta,k_l}\circ S_m(X)\right|,
\]
since only the first $m$ predicted output modes $\hat Y_{k_1},\dots, \hat Y_{k_m}$ can be compared to $Y_k$. We then consider the total loss $L_\theta$ of the approximation task, including the $N-m$ elements of the output $k$-space discarded by our model, i.e.
\[
    \begin{aligned}
        L_\theta(X, Y) &= \|Y - S^\top_m \hat Y\|= \underbrace{\sum_{l=0}^{m-1}\left|Y_{k_l} - \gamma_{\theta,k_l}\circ S_m(X)\right|}_{J_\theta(X, Y)} + \underbrace{\sum_{k=m}^{N-1}|Y_k - 0|}_{R_o(Y)}.
    \end{aligned}
\]
It follows that the overall loss $L_\theta$ is higher than \ourmethod{}'s training objective $J_\theta$, i.e. $L_\theta = J_\theta + R_o > J_\theta$,
% = \sum_{k=m}^{N-1}|Y_k|=\sum_{k=m}^{N-1}|\psi_k(X)|
whilst $R_o$ represents the \textit{irreducible} residual loss due to truncation of the predictions $\hat Y_k$. 
\paragraph{Optimal mode selection in auto-encoding \ourmethod{}} In case $Y = X$, i.e. the reduced-order \ourmethod{} is tasked with reconstructing the input spectrum, the optimal modes minimizing the irreducible loss are the ones with highest magnitude. This can be formalized as follows.
\begin{proposition}[Top-$m$ modes minimize the irreducible loss]
    Let $Y = X$ (reconstruction task). Then the choice ${k_0,\dots, k_{m-1}} = \topk_k{(m)}~~|X_{k}|$ minimizes the irreducible loss term $R_o$. %, i.e.

\end{proposition}
This means that if the spectrum of $X$ is monotonically decreasing in magnitude, then low-pass filtering is the optimal mode selection.
\begin{corollary}[Low pass filtering is optimal for monotonic spectrum] If $|X_k|$ is monotonically decreasing in $k$, then the choice $k_0,\dots,k_{m-1} = 0,\dots, m-1$ minimizes the residual $R_o$.
\end{corollary}
However, spectra in practical datasets are commonly non-monotonic e.g., the spectrum of solutions of chaotic or turbulent systems \citep{dumont1988characteristics}. We show an example in \cref{fig:era5}.

\paragraph{Mode selection criteria in general tasks} When $Y\neq X$ and the task is a general prediction task, the simple $\topk{m}$ analysis is not optimal. Nonetheless, given a dataset of input-output signals it is still possible to perform an \textit{a priori} analysis on $R_o$ to inform the choice of the best modes to keep. 

Often, we empirically observe the irreducible error $R_o$ for reduced-order \ourmethod{} to be smaller than for non-reduced-order FDMs i.e $R_o <\sum_{k=m}^{K-1}\|Y_k - \cT_k(\hat y)\|$ with layers of type \eqref{eq:1}\footnote{See \cref{fig:navier} and Appendix B for experimental evidence in support of this phenomenon.}.

We also note that the reachable component $J_\theta$ of the objective cannot always be minimized to zero regardless of the approximation power of $\gamma_{\theta}$. For each $k<m$, $S_m$ discards $N-m$ frequency components of the input signal which, if different than zero, likely contain the necessary information to approximate $\psi_k(X)$ exactly. Specifically, the irreducible lower bound on $J_\theta$ should depend on ``how much'' the output's $m$ frequency components depend on the discarded $N - m$ input's elements. 

A rough quantification of such bound can be obtained by inspecting the mismatch between the gradients of $\psi_k - \gamma_{\theta,k}\circ S_m$ with respect to $X$. In particular, it holds 
\[
    \sum_{j=0}^{N-1}\left|\frac{\partial\psi_k(X)}{\partial X_j} - \frac{\partial\gamma_{\theta,k}(S_m X)}{\partial X_j}\right| = \sum_{j=0}^{m-1}\left|\frac{\partial\psi_k(X)}{\partial X_j} - \frac{\partial\gamma_{\theta,k}(S_mX)}{\partial X_j}\right| + \sum_{j=m}^{N-1}\left|\frac{\partial\psi_k(X)}{\partial X_j}\right|,
\]
Unless $\partial_{X_j}\psi_k(X) = 0$ holds for all $j = m, \dots, N-1$ and $k = 0, 1, \dots, N-1$ i.e. no dependency of the ground truth map in $k$-space on the truncated elements, there will be an irreducible overall gradient mismatch and thus a nonzero $J_\theta$.

\subsection{Weight Initialization for Reduced-Order FDMs}\label{subsec:init}

FDMs \citep{li2020fourier,tran2021factorized,wen2022u} opt for a standard Xavier-like \citep{glorot2010understanding} initialization distribution that takes into account the input channels $c$ to a layer i.e. $\cN(0, \frac{1}{c})$. However, well-known variance preserving properties of Xavier schemes do not hold for FDM layers truncating $N - m$ elements of the $k$-space. Notably, Xavier schemes do not scale the variance of the weight initialization distribution based on the number of elements $m$ kept after truncation of the spectrum performed by $f_\theta$, leading to the \textit{collapse} of the outputs to zero.

To avoid this issue in \ourmethod{} and other FDMs, we develop a simple \textit{variance-preserving} (vp) that introduces a variance scaling factor based on $m$ and the class of transform.

{
\begin{restatable}[Variance Preserving ({\tt vp}) Initialization]{thm}{vpinit}\label{thm:vp}
Let $\hat x = W^* S_m^\top A S_m W x$ be a $k$-space reduced-order layer and $W$ is a normalized DCT-II transform. If $x\in\R^N$ is a random vector with 
\[
    \bE[x] = \0, \quad~ \mathbb{V}[x] = \sigma^2 \Id.
\]
Then,
\[
   A_{ij} \sim \mathcal{N}\left(0, \frac{{N}}{m^2}\right) \Rightarrow \mathbb{V}[\hat x] = \mathbb{V}[x].
\]
\end{restatable}
}

\begin{figure}[H]
    \vspace{-1mm}
    \includegraphics[width=\linewidth]{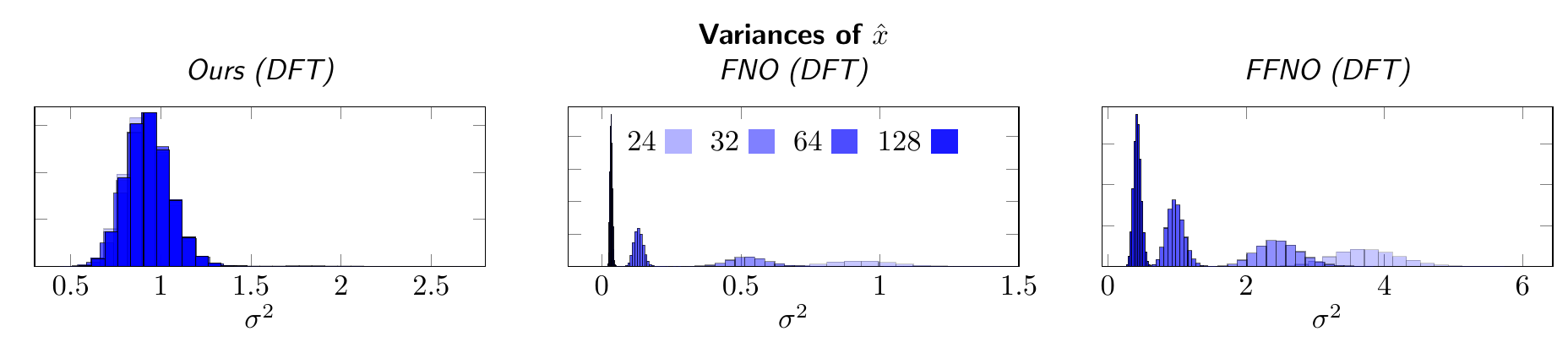}
    \vspace{-8mm}
    \caption{\small Output variance histogram in layer outputs $\hat x = W^*_mS_m^\top A(\theta) S_m W_N$, for a finite sample of inputs $x$ and a single sample of $\theta$. Color indicates signal resolution.}
    \label{fig:vpfig}
\end{figure}

We report the proof in Appendix A, including some considerations for specific forms of $f_\theta$.
{
\begin{restatable}[{\tt vp} initialization for DFTs]{corollary}{vpdft}\label{cor:vp_dft}
    Under the assumptions of Theorem \ref{thm:vp}, if $W$ is a normalized DFT matrix we have $\Re(A_{ij}),\Im(A_{ij}) \sim \mathcal{N}(0, \frac{N}{2m^2}) \Rightarrow \mathbb{V}[\hat x] = \mathbb{V}[x]$.
\end{restatable}
}
The collapse phenomenon is empirically shown in  \cref{fig:vpfig} for $m=24$, comparing a single layer of FNO and FFNO (with Xavier initialization) with FNO equipped with the proposed {\tt vp} scheme. Under the assumptions of Corollary \ref{cor:vp_dft}, we sample $A$ and compute empirical variances of $\hat x = W^*S_m^\top A(\theta)S_mWx$ for several finite batches of input signals $x$. We repeat the experiment for signals of different lengths $N$. The {\tt vp} scheme preserves unitary variances whereas the other layers concentrate output variances towards zero at a rate that grows with $N - m$.
\vspace{-6mm}
When the learned frequency-domain transformation $f_\theta$ is obtained, instead of the single low-rank linear layer $f_\theta = A(\theta) S_m X$, as the composition of several layers, preserving variances can be achieved by applying the {\tt vp} scheme only to the first layer. For some variants of FDMs e.g. FNO that truncate the spectrum at each layer, {\tt vp} initialization should instead be applied to all.

% Experiments
\clearpage\section{Experiments}\label{sec:experiments}
\begin{table}[b]
    \centering
    \begin{tabular}{c|ccc|cc}\toprule
        \textbf{Method} & Param. (M) & Size (MB) & Step (ms) & high $\nu$ & low $\nu$ \\
        \midrule
        FFNO \citep{tran2021factorized} & $8.9$ & $35$ & $294$ & $0.997${\small$\pm 0.003$} & $1.016${\small$\pm 0.010$}\\
        FNO \citep{li2020fourier} & $14.2$ & $56$ & $31$ & $0.379${\small$\pm 0.006$} & $0.328${\small$\pm 0.004$} \\
        FNOvp  & $14.2$ & $56$ & $32$ & $0.351${\small$\pm 0.003$} & $0.315${\small$\pm 0.006$} \\ 
        \rowcolor{blue!4}
        \ourmethod{+vp} & $10.2$ & $40$ & $19$ & $\textbf{0.257}${\small$\pm 0.007$} & $\textbf{0.240}${\small$\pm 0.004$} \\
    \end{tabular}
    \vspace{1mm}
    \caption{\small Benchmarks on incompressible Navier-Stokes. Direct long-range prediction errors (N-MSE) in $n$-space (signal space) of different models.}
    \label{tab:ns}
\end{table}
We validate \ourmethod{} on learning to approximate solution operators of dynamical systems from images. 
\begin{itemize}[leftmargin=5.5mm]
    \item In \cref{subsec:ns}, we apply \ourmethod{} on the standard task of learning solution operators for incompressible Navier-Stokes, comparing against other FDMs. In \cref{subsubsec:ns_ab} we perform a series of ablation experiments on each ingredient of the \ourmethod{} recipe, including weight initialization and architecture. In \cref{subsubsec:ns_sl} we provide scaling laws.
    \item In \cref{subsec:dfp} we deal with fluid-solid interaction dynamics in the form of higher resolution images ($128$). We consider turbulent flows around varying airfoil geometries, benchmarking against current SOTA \citep{thuerey2020deep}.
    \item In \cref{subsec:scal} we show how the computational efficiency of \ourmethod{} allows learning on unwieldy data without downsampling or building low-resolution meshes. We consider learning on high-resolution video ($600$ × $1062$) capturing the turbulent dynamics of smoke \citep{eckert2019scalarflow}.
\end{itemize} 

Configuration and model details are reported in the supplementary material. The code is available at \url{https://github.com/DiffEqML/kairos}. \textit{Weights \& Biases} ({\tt wandb}) \citep{wandb} logs of results are provided.

\subsection{Incompressible Navier-Stokes}\label{subsec:ns}
We show that \ourmethod{} matches or outperforms SOTA FDMs with less computation on the standard incompressible Navier-Stokes benchmark. Losses are reported in $n$-space (signal space) for comparison.

\paragraph{Setup}
We consider two-dimensional Navier-Stokes equations for incompressible fluid in vorticity form as described in \citep{li2020fourier}. Given a dataset of initial conditions, we train all models to approximate the solution operator at time $50$ seconds for high viscosity ($\nu = 1e^{-3}$) and at time $15$ for lower viscosity ($\nu = 1e^{-4}$). As a metric, we report \textit{normalized mean squared error} (N-MSE). Both initial condition as well as solution are provided as images of resolution $64$. 

We include as baseline established FDMs, such as Fourier Neural Operators (FNOs) \citep{li2020fourier} and \textit{Factorized Fourier Neural Operators} (FFNOs) \citep{tran2021factorized}. We indicate with the suffix \textit{vp} models that employ the proposed variance preserving initialization scheme. All models truncate to $m=24$, except FFNOs to $m=32$.

\begin{minipage}[t]{\linewidth}
    \centering
    \includegraphics[width=0.49\linewidth]{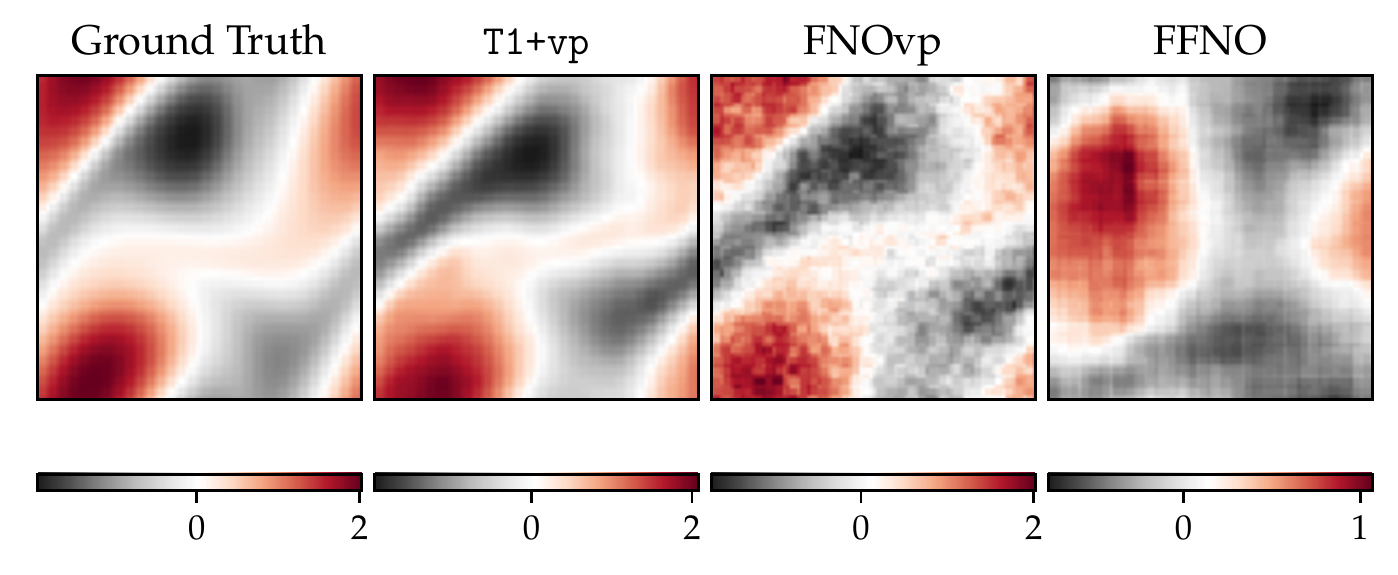}
    \hfill
    \includegraphics[width=0.49\linewidth]{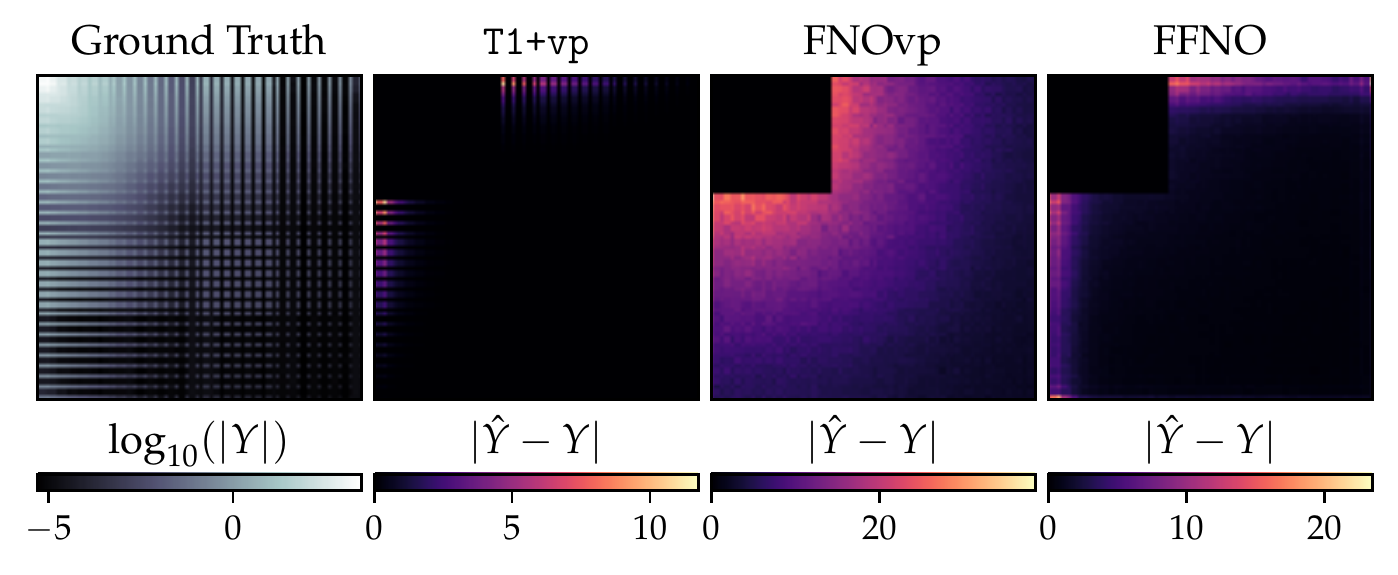}
    \vspace{-3mm}
    \captionof{figure}{\small \textbf{[Left]} Direct predictions at $T=50$s on high viscosity Navier-Stokes. \textbf{[Right]} Ground-truth spectrum and absolute errors in $k$-space (DCT-II). Despite predicting only the first $m=24$ elements, reduced-order \ourmethod{} models produce smaller errors even in other regions of the $k$-space.}
\label{fig:navier}
\end{minipage}

\paragraph{Results}
We perform $20$ training runs for each model and report mean and standard deviation in \cref{tab:ns}. \ourmethod{} reduces solution error w.r.t FNOs by over $20\%$ and FFNOs by over $40\%$. A single forward pass of \ourmethod{} models is on average $2$x faster than FNO and $10$x than FFNOs. We note that FFNOs are designed to share parameters between layers and thus require deeper architectures -- and slower, due to more transforms.  In particular, training time ($500$ epochs) for \ourmethod{} is cut to $20$ minutes down from $40$ of FNOs, matching the model speedup. Finally, we report an improvement in performance for FNOs with parameters initialized following our proposed scheme (FNOvp). \cref{fig:navier} provides sample predictions in $n$-space (left) to contextualize the task, in addition to prediction errors in frequency domain (right). Despite being a reduced order model with $m=24$, \ourmethod{+vp} produces smaller errors on truncated $k$-space elements ($k > m$) compared to FNOvp and FFNO.

\subsubsection{Ablations on weight scheme and architecture}\label{subsubsec:ns_ab}
\begin{wraptable}[9]{r}{0.35\linewidth}
    \vspace{-2mm}
    \centering
    \begin{tabular}{c|c|cc}\toprule
        \textbf{Method} & high $\nu$ & low $\nu$\\
        \midrule
        \ourmethod & $0.491$ & $0.449$\\ 
        \ourmethod{vp} & $0.304$ & $0.280$\\ 
        \ourmethod{+} & $0.295$ & $0.260$\\
        \rowcolor{blue!4}
        \ourmethod{+vp} & $\textbf{0.257}$ & $\textbf{0.240}$\\
    \end{tabular}
    \vspace{-1.4mm}
    \caption{\small Ablation on the effect of the proposed weight initialization scheme and \ourmethod{} architecture.}
    \label{tab:ablans}
\end{wraptable}

We repeat the previous experiment and report prediction errors for four variants of \ourmethod{}: same architecture and weight initialization scheme as FNOs (\ourmethod{}), \ourmethod{} with our proposed vp scheme (\ourmethod{vp}), a reduced-order variant with $k$-space model $f_\theta$ defined as a UNet architecture (\ourmethod{+}), and \ourmethod{+} with variance preserving scheme (\ourmethod{+vp}). The results in \cref{tab:ablans} provide empirical evidence in support of the vp scheme and its synergistic effect with the proposed architecture. In particular, combining vp scheme and UNet structure in frequency domain reduces error by half compared to the naive \ourmethod{} approach.

\subsubsection{Scaling laws}\label{subsubsec:ns_sl}
\begin{wrapfigure}[9]{r}{0.4\linewidth}
    \centering
    \vspace{-10mm}
    \includegraphics[width=\linewidth]{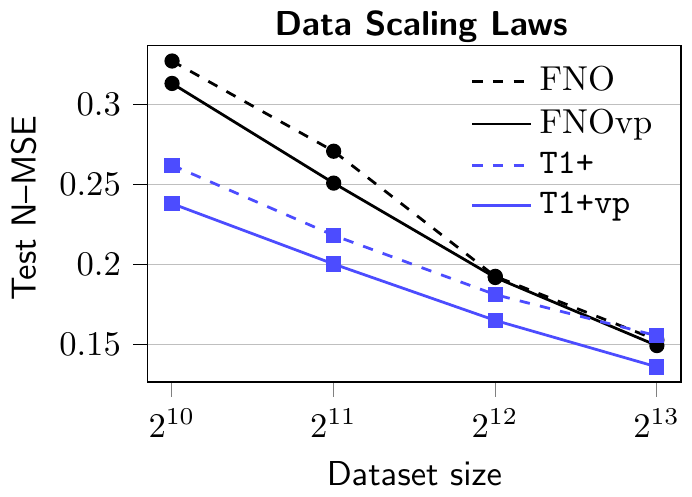}
    \vspace{-5.5mm}
    \caption{\small Scaling laws for N-MSE.}
    \label{fig:scaling_law}
\end{wrapfigure}
We verify whether the reduction in predictive error of \ourmethod{} over neural operator baselines is preserved as the size of training dataset grows. We perform $10$ training runs on the Navier-Stokes $\nu=1e^{-4}$ experiment, each time with a larger dataset size, and report the scaling laws in \cref{fig:scaling_law}. With additional data, the gaps in test errors narrow slightly, with noticeable improvements obtained by applying the {\tt vp} scheme to both FNO and {\tt T1+}.

\subsection{Flow Around Airfoils}\label{subsec:dfp}
We investigate the performance of \ourmethod{} in predicting steady-state solutions of flow around airfoils. 

\paragraph{Setup}

We use data introduced in \citep{thuerey2020deep} in the form of $10000$ training pairs of initial conditions, specifying freestream velocities and the airfoil mask, with the target steady-state velocity and pressure fields. This task introduces additional complexity in the form of higher resolution input images ($128$) and a full $k$-space due to the discontinuity in the field produced by the mask. 

We compare a SOTA UNet architecture (DFPNet) introduced by \citep{thuerey2020deep} to FNOs and \ourmethod{} with {\tt vp} initialization schemes. We perform a search on the most representative hyperparameters (detailed in the Appendix). Averages for $5$ runs are reported in \cref{tab:dfp}.

\paragraph{Results}

\begin{wraptable}[9]{l}{0.43\linewidth}
    \vspace{-2mm}
    \centering
    \begin{tabular}{c|c|c}\toprule
        \textbf{Method} & N-MSE & Time (hrs) \\
        \midrule
        DFPNet & $0.023$ & $\textbf{1.3}$\\ 
        FNO & $\textbf{0.020}$ & $6.0$\\ 
        \rowcolor{blue!4}
        \ourmethod{+}vp & $0.024$ & $\textbf{1.3}$\\
    \end{tabular}
    \vspace{-0.2mm}
    \caption{\small Test N-MSE and total training time on the flow around airfoil task.}
    \label{tab:dfp}
\end{wraptable}

All models are able to accurately predict steady-state solutions for different airfoils with small normalized errors. Test N-MSE is comparable as all models are within a single standard deviation. Training of \ourmethod{} is as fast as DFPNets \citep{thuerey2020deep} and as accurate as FNOs, as evidence of the applicability of \ourmethod{} to tasks with signals that are not band-limited (in this case due to the airfoil mask).

\clearpage

\subsection{Turbulent Smoke}\label{subsec:scal}
We investigate the performance of \ourmethod{} in predicting iterative rollouts from high-resolution video of real rising smoke plumes.

\begin{minipage}[t]{\linewidth}
    \centering
    \includegraphics[width=0.49\linewidth]{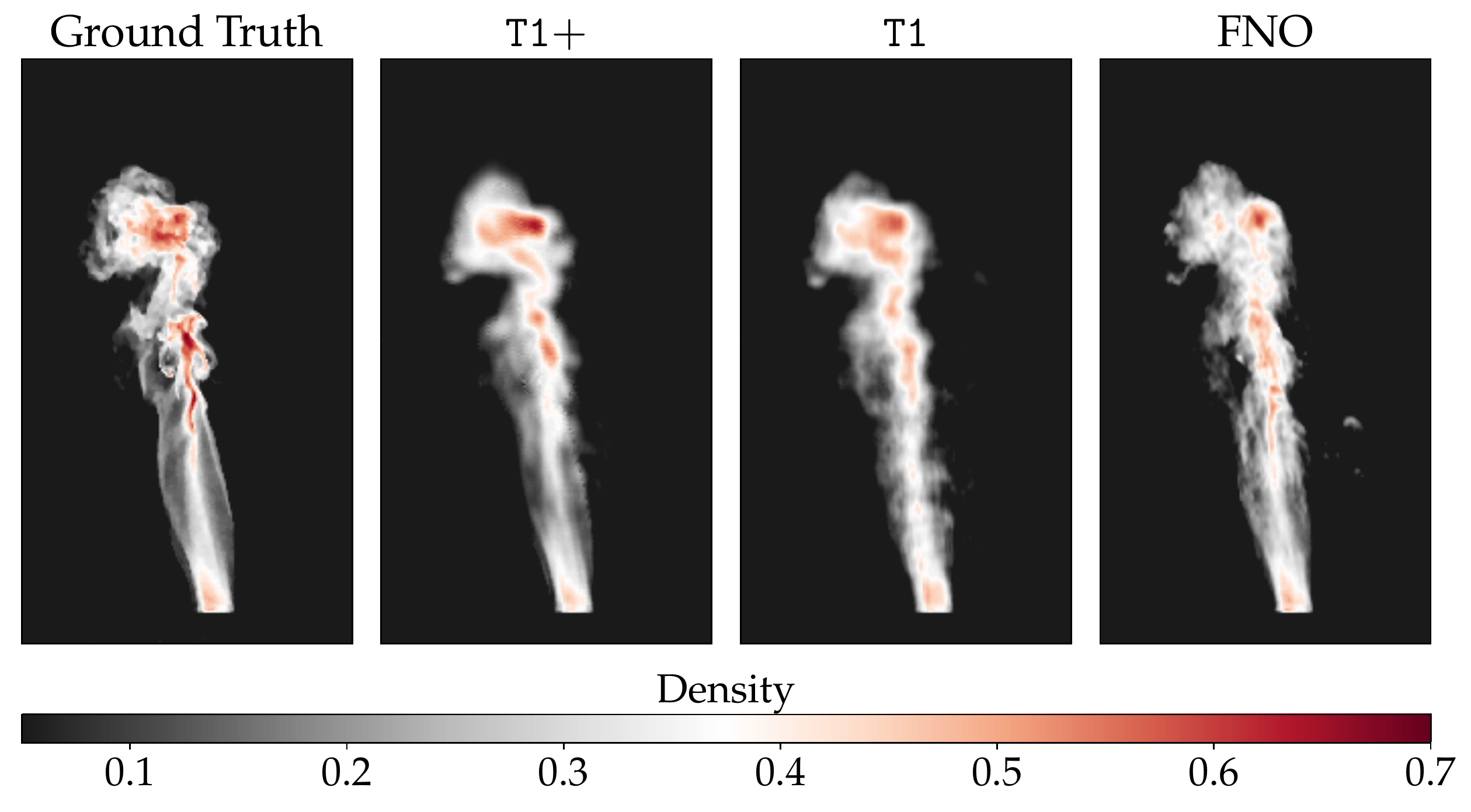}
    \hfill
    \includegraphics[width=0.49\linewidth]{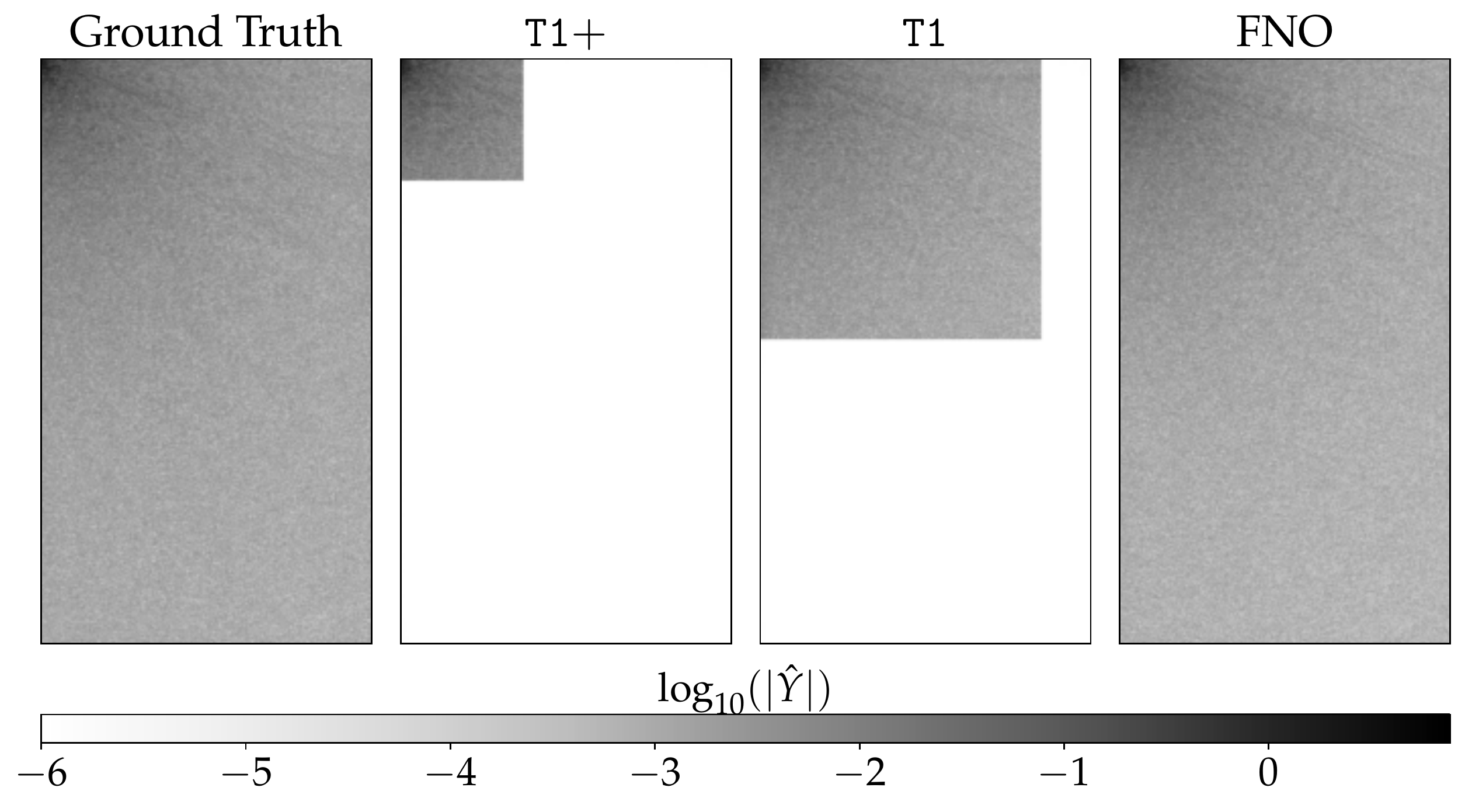}
    \vspace{-3mm}
     \captionof{figure}{\small \textbf{[Left]} $10$-step rollout predictions on ScalarFlow. FNOs produce high-frequency, non-physical artifacts and accumulate error more rapidly in time compared to \ourmethod{} models \textbf{[Right]} Log-absolute values of predictions in $k$-space (DCT-II). Although \ourmethod{} is limited to $m=512$ and \ourmethod{+} to $m=224$ $k$-space elements, the predictions are overall more physically accurate in $n$-space.}
\label{fig:scalarflow-comparison-raw}
\end{minipage}
% \hfill

\paragraph{Setup}
We use the ScalarFlow dataset introduced in \citep{eckert2019scalarflow} consisting of $104$ sequences of $150$ frames each collected from video recordings of rising hot smoke plumes. The dataset consists of raw video data at high-resolution ($600$ × $1062$) collected at $60$ fps. This task scales up complexity by involving real-world high-definition data, capturing highly-turbulent dynamics. We perform rollouts iteratively based on previous predictions: all models are trained on $3$-step rollouts and evaluated over $10$-steps extrapolation to test their generalization in time. We compare FNOs against \ourmethod{}, \ourmethod{+} and \ourmethod{+vp} of similar model sizes after performing a search on most representative hyperparameters (Appendix B).

%  FNO models truncate to $m=48$ with residual connections, \ourmethod{} to $m=512$ and \ourmethod{+} to $m=224$.
 
 \begin{wraptable}[9]{r}{0.42\linewidth}
    \vspace{-3.6mm}
    \centering
    \begin{tabular}{c|c|c}\toprule
        \textbf{Method} & N-MSE & Time (hrs) \\
        \midrule
        FNO & $0.232$ & $32.4$\\ 
        \ourmethod{} & $0.239$ & $8.1$\\ 
        \ourmethod{+} & $0.256$ & $4.7$\\
        \rowcolor{blue!4}
        \ourmethod{+vp} & $\textbf{0.228}$ & $\textbf{4.7}$\\
    \end{tabular}
    \vspace{-1.1mm}
    \caption{\small Test 10-steps rollout $n$-space prediction errors (N-MSE)  and total training time on the ScalarFlow dataset.}
    \label{tab:scalarflow-small}
\end{wraptable}

\paragraph{Results}
\cref{fig:scalarflow-comparison-raw} provides a sample rollout of different model predictions in $k$-space (DCT-II). \ourmethod{+vp} accumulates smaller errors over the rollout and is less prone to generating non-physical artifacts by performing prediction only on a subset of the $k$-space (\cref{tab:scalarflow-small}). Notably, \ourmethod{} and \ourmethod{+} are $4\times$ to $7\times$ faster, providing a reduction in training time from $32.4$ hours to $4.7$. Appendix B includes additional visualizations, including averaged prediction errors on $k$-space.

\section{Conclusion}

We present a streamlined class of \textit{frequency domain models} (FDM): \textit{Transform Once} (\ourmethod{}). \ourmethod{} models are optimized directly in frequency domain, after a single transform, and achieve similar or improved predictive performance at a fraction of the computational cost ($3$x to $10$x speedups across tasks). Further, a simple truncation-aware weight initialization scheme is introduced and shown to improve the performance of \ourmethod{} and existing FDMs.

\section*{Acknowledgments}
This work is supported by NSF (1651565), AFOSR (FA95501910024), ARO (W911NF-21-1-0125), ONR, DOE, CZ Biohub, Sloan Fellowship and JSPS Kakenhi (21J14546).

%%%%% BACK MATTER %%%%%
%%%%% BACK MATTER %%%%%

% Bibliography
\bibliographystyle{abbrvnat}
\bibliography{main_final.bib}

% Checklist
%\input{sections/back_matter/checklist}

% Acknowledgements
%\input{sections/back_matter/acknowledgements}

% Funding
%\input{sections/back_matter/funding}

%%%%% APPENDIX %%%%%
% Table of Contents

\newpage
\begin{center}
    \huge{\bf{\textit{Transform Once}} \\
    \emph{Supplementary Material}}
\end{center}
\vspace*{3mm}

\appendix
\addcontentsline{toc}{section}{}
\part{}
\parttoc

% Notation
\paragraph{Notation}

We report here a reference for notation used in main text and supplementary.
\begin{table}[H]
    \centering
    \begin{tabular}{c|l}\toprule
        Symbol & Description \\\midrule
        $\R$ & Set of reals \\
        $\bC$ & Set of complex numbers\\
        $\bE[x]$ & Expected value of random variable $x$\\
        $\bV[x]$ & Variance of random variable $x$\\
        $\Sigma_x$ & Covariance matrix of random variable $x$\\
        $\tr$ & Trace operator for square matrices. $\tr(A) = \sum_{n}A_{nn}$\\
        $\circ$ & Composition of functions $f\circ g(x) = f(g(x))$\\  
        $*$ & Conjugate transpose operator. $A^* = \bar A^\top$ where $\bar A$ has complex conjugated entries \\
        $\wedge$ & Outer product $u\wedge v = uv^*$ for $u,v\in\bC^n$\\
        \bottomrule
    \end{tabular}
\end{table}

\clearpage

% Derivations and Background
\section{Proof of Theorem \ref{thm:vp}}
\subsection{Preliminary Lemmas}
\begin{lemma}[Propagation of Uncertainty under DFT/DCT] \label{lem:A1}
    Let $X = Wx$ with $x\in\mathbb R^N$ and $W\in\mathbb C^{N\times N}$. Then
    \[
        \Sigma_X = W \Sigma_x W^*
    \]
\end{lemma}
\proof
    \[
        \begin{aligned}
            \Sigma_X &= \mathbb E\left[ (Wx - \mathbb{E}[Wx])\wedge (Wx - \mathbb{E}[Wx])\right]\\
                     &= \mathbb E\left[ W(x - \mathbb{E}[x])\wedge W(x - \mathbb{E}[x])\right]\\
                     &= \mathbb E\left[ W(x - \mathbb{E}[x])(x - \mathbb{E}[x])^\top W^*\right]\\
                     &= W\mathbb E\left[ W(x - \mathbb{E}[x])(x - \mathbb{E}[x])^\top \right]W^*\\
                     &= W \Sigma_x W^*
        \end{aligned}
    \]
\endproof
\begin{lemma}[Propagation of Total Variance under DFT/DCT] \label{lem:A2} Let $X = Wx$ with $x\in\mathbb R^N$ and $W\in\mathbb C^{N\times N}$. Then
$$
    \mathbb V[X] = \mathbb{V}[x]
$$
\end{lemma}
\proof
    Recalling that the total variance of a random variable is equal to the trace of its covariance matrix, i.e. 
    \[
        \mathbb{V}[x] = \text{tr}(\Sigma_x),\quad\mathbb{V}[X] = \text{tr}(\Sigma_X)
    \]
    then
    \[
        \begin{aligned}
                            \text{tr}(\Sigma_x) =  \text{tr}(\Sigma_X) \Leftrightarrow  \mathbb V[X] = \mathbb{V}[x]
\        \end{aligned}
    \]
    
    Recalling Lemma \ref{lem:A1} yields
    \[
        \begin{aligned}
                                  & \mathbb V[X] = \mathbb{V}[x]\\
            \Leftrightarrow \quad & \text{tr}(\Sigma_x) =  \text{tr}(W\Sigma_x W^*)\\
            \Leftrightarrow \quad & \text{tr}(\Sigma_x) -  \text{tr}(W\Sigma_x W^*) = 0\\
            \Leftrightarrow \quad & \text{tr}(\Sigma_x) -  \text{tr}(\Sigma_x W^*W) = 0\\
            %\Leftrightarrow \quad & \sum_{n}[\Sigma_x]_{nn} -  \sum_{n,k,j}[W]_{nk}[\Sigma_x]_{kj}[W^*]_{jn} = 0\\
            %\Leftrightarrow \quad & \sum_{n}[\Sigma_x]_{nn} -  \sum_{n}[\Sigma_x]_{nn}[W]_{nk}[\Sigma_x]_{kj}[W^*]_{jn} = 0
        \end{aligned}
    \]
    Since the DCT/DFT matrix is orthonormal, i.e. $W^* = W^{-1}$ we have that 
    \[
        \text{tr}(\Sigma_x W^*W) = \text{tr}(\Sigma_x),
    \]
    proving the result.
\endproof
\begin{lemma}[Gaussian initialization in rank--deficient linear layers] \label{lem:A3} Let $\hat X = S^\top_m A S_m X$ with $X\in\R^N$, $A\in\bC^{m\x m}$ and $S_m\in\bC^{m\times N}$,

% \[
% {S_m} = 
% \begin{tikzpicture}[baseline=(current  bounding  box.center),mymatrixenv]
%     \matrix [mymatrix,inner sep=4pt] (m)  
%     {
% \tikzmarkin[kwad=style green]{Prime} 1 & \cdots & 0 & \tikzmarkin[kwad=style cyan]{Bis} 0 & \tikzmarkend{Bis} \cdots & \tikzmarkend{Bis} 0 \\
% \vdots  & \ddots & \vdots & \tikzmarkend{Bis} \vdots & \tikzmarkend{Bis} \ddots & \tikzmarkend{Bis} \vdots  \\
% 0 & \cdots &  1 \tikzmarkend{Prime} & \tikzmarkend{Bis} 0 & \tikzmarkend{Bis} \cdots & 0 \tikzmarkend{Bis}  \\    
% };
% % Braces     
% \mymatrixbraceright{1}{3}{$m$}
% \mymatrixbracetop{1}{3}{$m$}
% \mymatrixbracetop{4}{6}{$T - m$}
% \end{tikzpicture}
% \begin{tikzpicture}[baseline=(current  bounding  box.center)]
%         \matrix (m) [matrix of math nodes,row sep=3em,column sep=4em,minimum width=2em]
%         {
%          x & \hat x \\
%          X & \hat X \\};
%         \path[-stealth]
%         (m-1-1) edge node [left] {$W$} (m-2-1)
%         (m-2-1.east|-m-2-2) edge node [below] {$S_m^\top A(\theta) S_m$} (m-2-2)
%         (m-2-2) edge node [right] {$W^*$} (m-1-2);
%     \end{tikzpicture}
% \]
\pgfkeys{tikz/mymatrixenv/.style={decoration={brace},every left delimiter/.style={xshift=8pt},every right delimiter/.style={xshift=-8pt}}}
\pgfkeys{tikz/mymatrix/.style={matrix of math nodes,nodes in empty cells,left delimiter={[},right delimiter={]},inner sep=1pt,outer sep=1.5pt,column sep=8pt,row sep=8pt,nodes={minimum width=20pt,minimum height=10pt,anchor=center,inner sep=0pt,outer sep=0pt}}}
\pgfkeys{tikz/mymatrixbrace/.style={decorate,thick}}

\newcommand*\mymatrixbraceright[4][m]{
    \draw[mymatrixbrace] (#1.west|-#1-#3-1.south west) -- node[left=2pt] {#4} (#1.west|-#1-#2-1.north west);
}
\newcommand*\mymatrixbraceleft[4][m]{
    \draw[mymatrixbrace] (#1.east|-#1-#2-1.north east) -- node[right=2pt] {#4} (#1.east|-#1-#2-1.south east);
}
\newcommand*\mymatrixbracetop[4][m]{
    \draw[mymatrixbrace] (#1.north-|#1-1-#2.north west) -- node[above=2pt] {#4} (#1.north-|#1-1-#3.north east);
}
\newcommand*\mymatrixbracebottom[4][m]{
    \draw[mymatrixbrace] (#1.south-|#1-1-#2.north east) -- node[below=2pt] {#4} (#1.south-|#1-1-#3.north west);
}
\tikzset{style green/.style={
    set fill color=green!50!lime!60,draw opacity=0.4,
    set border color=green!50!lime!60,fill opacity=0.1,
  },
  style cyan/.style={
    set fill color=cyan!90!blue!60, draw opacity=0.4,
    set border color=blue!70!cyan!30,fill opacity=0.1,
  },
  style orange/.style={
    set fill color=orange!90, draw opacity=0.8,
    set border color=orange!90, fill opacity=0.3,
  },
  style brown/.style={
    set fill color=brown!70!orange!40, draw opacity=0.4,
    set border color=brown, fill opacity=0.3,
  },
  style purple/.style={
    set fill color=violet!90!pink!20, draw opacity=0.5,
    set border color=violet, fill opacity=0.3,    
  },
  kwad/.style={
    above left offset={-0.1,0.23},
    below right offset={0.10,-0.36},
    #1
  },
  pion/.style={
    above left offset={-0.07,0.2},
    below right offset={0.07,-0.32},
    #1
  },
  poz/.style={
    above left offset={-0.03,0.18},
    below right offset={0.03,-0.3},
    #1
  },set fill color/.code={\pgfkeysalso{fill=#1}},
  set border color/.style={draw=#1}
}
\vspace*{-6mm}
\[
    {S_m} = 
    \begin{tikzpicture}[baseline={-0.5ex},mymatrixenv]
        \matrix [mymatrix,inner sep=4pt] (m)  
        {
    \tikzmarkin[kwad=style green]{Prime} 1 & \cdots & 0 & \tikzmarkin[kwad=style cyan]{Bis} 0 & \tikzmarkend{Bis} \cdots & \tikzmarkend{Bis} 0 \\
    \vdots  & \ddots & \vdots & \tikzmarkend{Bis} \vdots & \tikzmarkend{Bis} \ddots & \tikzmarkend{Bis} \vdots  \\
    0 & \cdots &  1 \tikzmarkend{Prime} & \tikzmarkend{Bis} 0 & \tikzmarkend{Bis} \cdots & 0 \tikzmarkend{Bis}  \\    
    };
    % Braces     
    \mymatrixbraceright{1}{3}{\small$m$}
    \mymatrixbracetop{1}{3}{\small$m$}
    \mymatrixbracetop{4}{6}{\small$N - m$}
    \end{tikzpicture}.
\]
\vspace*{1mm}

If $\bE[X_k] = 0$, $\bV[X_k] = \sigma^2$ for all $k$ the following hold:
%and $p_{\Re(A_{ij})} = p_{\Im(A_{ij})} = \cN(0, \sigma_A)$ for all entries of $A$, then 
%
\begin{itemize}
    \item[$i.$] for $k\geq m$
    \[
        \bE[\hat X_k] = 0,\quad
        \bV[\hat X_k] = 0
    \]
    \item[$ii.$] for $k<m$ and $\Re(A_{ij}),\Im(A_{ij})\sim \cN(0, \sigma_A^2)$
    \[
        \bE[\hat X_k] = 0,\quad
        \bV[\hat X_k] = 2 m \sigma^2 \sigma_A^2
    \]
    \item[$iii.$] for $k<m$ and $\Re(A_{ij})\sim \cN(0, \sigma_A^2)$, $\Im(A_{ij})=0$
    \[
        \bE[\hat X_k] = 0,\quad
        \bV[\hat X_k] = m \sigma^2 \sigma_A^2
    \]
\end{itemize}
\end{lemma}
\proof
    Let $M = S_m^\top A S_m$. It holds,
    \[
        M = 
            \begin{bmatrix}
                A & \x\\
                \x & \x
            \end{bmatrix}\in\bC^{N\x N}
    \]
    where``$\x$'' are blocks of complex zeros. By expanding component--wise the layer computation, i.e.
    \[
        \hat X_k = \sum_{j=0}^{N-1} M_{kj}X_j,
    \]
    it holds that for $k<m$
    \[
        \hat X_k = \sum_{j=0}^{m-1} A_{kj}X_j,
    \]
    while $\hat X_k = 0$ for $k\geq m$. Hence $i.$ follows naturally from the latter and we focus on proving $ii.$ and $iii.$
    \begin{itemize}
        \item[Case $ii.$] The probability distribution of $\hat X_k$ is a sum of product distributions involving independent random variables $A_{kj}$ and $X_j$. The first central moment is readily obtained
        \[
            \mathbb{E}[\hat{X}_k] = \sum_{t=0}^{m-1} \mathbb{E}[A_{kj}] \mathbb{E}[X_j] = 0
        \]
    since both $\mathbb{E}[X_k] = 0$ and $\forall~ k,j<m: ~ \mathbb{E}[A_{kj}] = 0$. $\bV[\hat X_k]$ can be then obtained by computing the variance of the product of two random variables, i.e.
    \begin{equation*}
        \begin{aligned}
            \mathbb{V}[\hat X_k] &= \sum_{j=0}^{m-1} \Big(\mathbb{V}[A_{kj}] + \cancel{\mathbb{E}[A_{kj}]}^2) (\mathbb{V}[{X_{j}}] + \cancel{\mathbb{E}[{X_{j}}]}^2) - \cancel{\mathbb{E}[A_{kj}]^2\mathbb{E}[{X_{j}}]^2}\Big) \\
            &= \sum_{j=0}^{m-1} \mathbb{V}[A_{kj}] \mathbb{V}[{X_{j}}] \\
            &= \sum_{j=0}^{m-1} \sigma^2 \mathbb{V}[A_{kj}] \\
            &= \sigma^2\sum_{j=0}^{m-1}\left(\mathbb{V}[\Re(A_{kj})] + \mathbb{V}[\Im(A_{kj})]\right)\\
            &= \sigma^2\sum_{j=0}^{m-1}2\sigma^2_A =  2 m \sigma^2 \sigma_{A}^2
        \end{aligned}
    \end{equation*}
    \item[Case $iii.$] Similarly to the previous case we get 
    \[
        \begin{aligned}
            \mathbb{V}[\hat X_k] &= \sigma^2\sum_{j=0}^{m-1}\left(\mathbb{V}[\Re(A_{kj})] + \cancel{\mathbb{V}[\Im(A_{kj})}]\right)\\
            &= \sigma^2\sum_{j=0}^{m-1}\sigma^2_A=  m \sigma^2 \sigma_{A}^2
        \end{aligned}
    \]
    \end{itemize}
\endproof
\subsection{Proof of Main Result}

\proof
    According to Lemma \ref{lem:A2}, the total variance is preserved under the normalized DCT. Therefore, with $X = W\hat x$ and $\hat X = Wx$ we have
    \[
        \bV[X] = \bV[x], \quad \bV[\hat X] = \bV[\hat x].
    \]
    Using $\hat X = S_m^\top A S_m X$, we can find the condition under which the variance is preserved by the map $x\mapsto \hat x$:
    \[
        \begin{aligned}
                             & \mathbb{V}[\hat x] =  \mathbb{V}[x]\\
        \Leftrightarrow\quad & \sum_{n=0}^{N-1} \mathbb{V}[\hat x_n] = \sum_{n=0}^{N-1} \mathbb{V}[x_n]\\
        \Leftrightarrow\quad & \sum_{k=0}^{N-1} \mathbb{V}[\hat X_k] = \sum_{k=0}^{N-1} \mathbb{V}[X_k]\\
        \Leftrightarrow\quad & \sum_{k=0}^{m-1} m\sigma^2\sigma^2_A = \sum_{k=0}^{N-1} \sigma^2&&\quad \text{Lemma \ref{lem:A3}}\\
        \Leftrightarrow\quad & m^2\sigma^2\sigma^2_A =  N \sigma^2 \\
        \Leftrightarrow\quad & \sigma^2_A =  \frac{N}{m^2}
        \end{aligned}
    \]
    Hence, initializing $A$ by sampling its entries from a normal distribution with zero mean and variance $N/m^2$ is sufficient for preserving the variance under the reduced-order FDM layer, i.e.
    \[
        A_{ij} \sim \cN\left(0, \frac{N}{m^2}\right)~~\Rightarrow~~\bV[\hat x] = \bV[x],
    \]
    proving the result.
\endproof

\vpdft*
\proof
    The proof follows directly from the one of Theorem \ref{thm:vp} using the fact that since the DFT's $k$-space is complex ($\cD_k\equiv\bC^N$) as $W\in\bC^{N\times N}$, the weights are typically chosen complex, i.e. $A\in\bC^{m\times m}$. Therefore, in this case $\bV[\hat X] = 2m\sigma^2\sigma_A^2$ according to Lemma \ref{lem:A3}.
\endproof

\begin{restatable}[({\tt vp}) initialization with diagonal layers]{corollary}{vpinit_diag}\label{cor:vp_diag}
Under the assumptions of Theorem \ref{thm:vp}, if $A$ is diagonal s.t $\forall i \neq j: A_{ij} = 0$, we have
$
    A_{ii} \sim \cN\left(0, \frac{N}{m}\right)~~\Rightarrow~~\bV[\hat x] = \bV[x].
$
    
\end{restatable}
\proof
The proof follows directly from Lemma \ref{lem:A3}
\begin{equation*}
    \begin{aligned}
        \mathbb{V}[\hat X_k] &= 
        \sum_{j=0}^{m-1} \mathbb{V}[A_{kj}] \mathbb{V}[{X_{j}}] \\
        &= \mathbb{V}[A_{kk}] \mathbb{V}[{X_{k}}] \\
        &= \sigma^2 \left(\mathbb{V}[\Re(A_{kk})] + \cancel{\mathbb{V}[\Im(A_{kk})]}\right)\\
        &= \sigma^2\sigma^2_A 
    \end{aligned}
\end{equation*}
leading to the condition 
    \[
        \begin{aligned}
         & \mathbb{V}[\hat x] =  \mathbb{V}[x]\\
        \Leftrightarrow\quad & \sum_{k=0}^{m-1} \sigma^2\sigma^2_A = \sum_{k=0}^{N-1} \sigma^2\\
        \Leftrightarrow\quad & m\sigma^2\sigma^2_A =  N \sigma^2 \\
        \Leftrightarrow\quad & \sigma^2_A =  \frac{N}{m}
        \end{aligned}
    \]
\endproof

The layer structure treated by \ref{cor:vp_diag} is common among many FDMs, e.g. FNOs in \citep{li2020fourier}.
% Experimental Details

\section{Additional Details}
\paragraph{Broader impact}
FDMs are widely used in the context of learning to predict the evolution of dynamical systems. The model class presented in this work, \ourmethod{}, provides an accessible way to train and evaluate large-scale FDMs, reducing memory overhead and overall training times. When predicting the solution of e.g. a \textit{partial differential equation} (PDE), care should be taken especially when the prediction is used to inform downstream decision making, as many systems are optimally predictable only for a certain time scale \citep[pp. 366]{strogatz2018nonlinear}. We anticipate a potential positive environmental impact from the adoption of \ourmethod{} as a replacement for the largest FDMs currently in use. 

\paragraph{Experimental setup} Experiments have been performed on an \textsc{NVIDIA\copyright~DGX} workstation equipped with a 128 threads \textsc{AMD\copyright~Epyc 7742} CPU, 512GB of RAM and four \textsc{NVIDIA\copyright~A100} GPUs. The main software implementation has been done within the $\tt PyTorch$ \citep{paszke2017automatic} ecosystem building upon the $\tt pytorch$-$\tt lightning$ \citep{falcon2019pytorch} framework. 

\paragraph{Common experimental settings} 
\subsection{Incompressible Navier–Stokes}\label{asec:exp_nvs}
\paragraph{Dataset}
We use data generated in \citep{li2020fourier} in the form of pairs of initial conditions and solutions of the incompressible Navier-Stokes equations in vorticity form solved with a pseudospectral method. The dataset \footnote{Data can be downloaded here: \href{https://drive.google.com/drive/folders/1UnbQh2WWc6knEHbLn-ZaXrKUZhp7pjt-}{Google Drive link}. High viscosity: {\tt NavierStokes\_V1e-3\_N5000\_T50}, Low viscosity: {\tt NavierStokes\_V1e-4\_N10000\_T30}.} is comprised rollouts of solutions as images of resolution $64$.

\paragraph{Models and training}
The training configuration is shared by all models:

\begin{listing}[H]
\begin{mintedbox}{yaml}
datamodule: 
    ntrain: 1000
    ntest: 200
    batch_size: 64
    history_size: 1
train:
    optimizer: 
        type: AdamW
        learning_rate: 1e-3
        weight_decay: 1e-4
    scheduler: 
        type: Step
        step_size: 100
        gamma: 0.5
        scheduler_interval: epoch
loss_fn: RelativeL2Loss
\end{mintedbox}
\vspace{-6mm}
\end{listing}
For the high viscosity ($1e^{-3}$) setting, the models are trained to predict the solution at time $T=50$ seconds directly, without producing rollouts and supervising the model with solutions at times between $0$ and $50$. Crucially, this ensures that the task is much more challenging than that of \citep{li2020fourier}, where for a single training sample the entire rollout is used as supervision. For the low viscosity setting ($1e^{-4}$), target times are $T=15$ seconds.

Model configurations are given below:

\begin{listing}[H]
\begin{minipage}[t]{0.32\textwidth}
\begin{mintedbox}{yaml, title=\config{FNO}}
FNO:
    modes: 24
    nlayers: 6
    width: 32 \end{mintedbox}
\end{minipage}
\begin{minipage}[t]{0.32\textwidth}
\begin{mintedbox}{yaml, title=\config{\ourmethod{}}}
T1:
    modes: 24
    nlayers: 6
    width: 48  \end{mintedbox}
\end{minipage}
\begin{minipage}[t]{0.32\textwidth}
\begin{mintedbox}{yaml, title=\config{FFNO}}
FFNO:
    modes: 32
    nlayers: 10
    width: 82 \end{mintedbox}
\end{minipage}
\vspace{-6mm}
\end{listing}

where each layer in a model shares the same structure. In FNOs and FFNOs, we employ a regular FDM layer following \citep{li2020fourier,tran2021factorized} with k-space convolutions and residual connections given by n-space layers (pointwise convolutions for FNOs, dense for FFNOs). \ourmethod{} uses a similar layer without n-space residual paths. The differences in number of layers and width have been introduced to keep parameter counts comparable. At a given channel width, FNOs require the largest number of parameters due to k-space convolutions on complex numbers given by the DFT coefficients. Although FFNOs \citep{tran2021factorized} are most parameter efficient due to parameter sharing, we found them unable to tackle the task and produce high-quality predictions.  

\begin{figure*}[h!]
    \centering
    \includegraphics[width=0.8\linewidth]{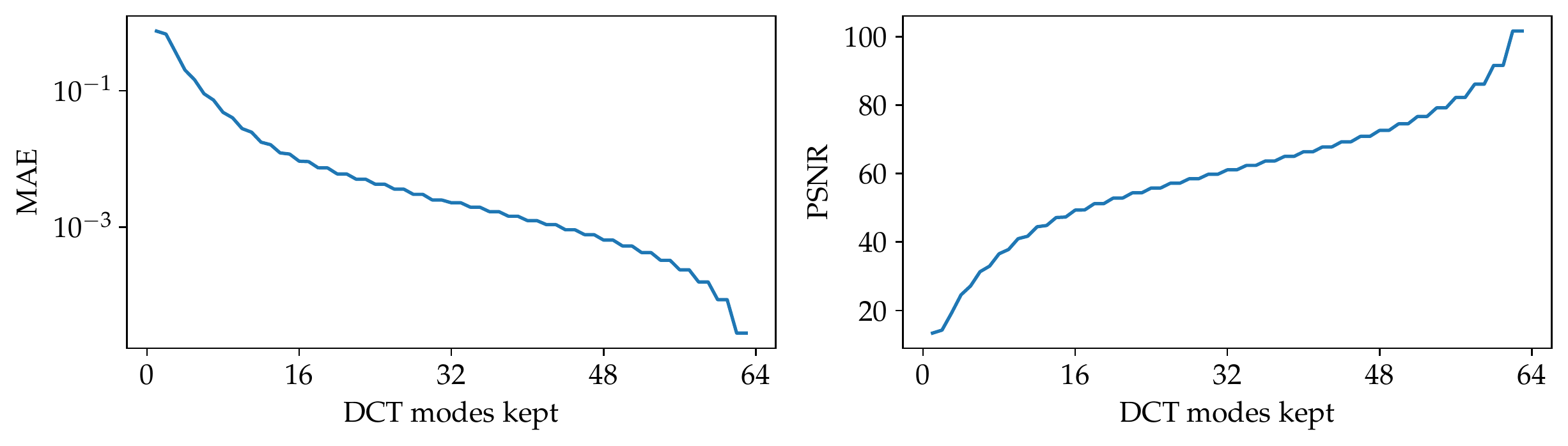}
    \vspace{-3mm}
    \caption{\small Incompressible Navier-Stokes: metrics vs number of DCT modes (i.e. $m$ elements) kept (i.e. not pruned).}
    \label{fig:navier-stokes-modes-mae-psnr}
\end{figure*}

\ourmethod{+} employs a UNet on the patch constructed by the elements of the k-space kept, and shares its structure with \ourmethod{} otherwise. The {\tt vp} parameter initialization scheme in \ourmethod{} is applied only to the first layer performing the truncation in k-space, not to the following layers which use standard Kaiming initialization \cite{he2015delving}. In FNOvp the scheme is applied to all layers.

\begin{figure*}[h!]
    \centering
    \includegraphics[width=\linewidth]{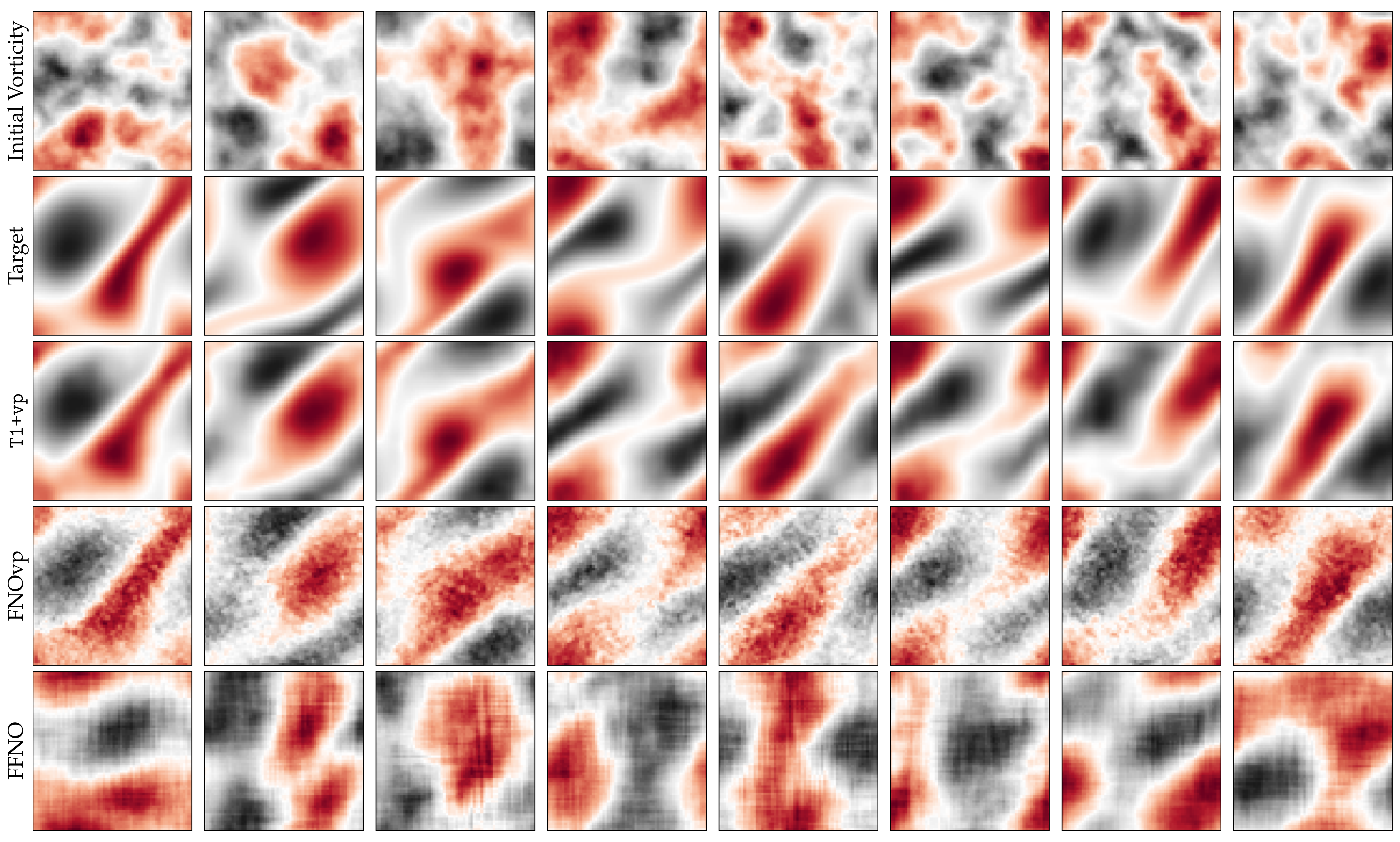}
    \vspace{-3mm}
    \captionof{figure}{\small Initial conditions, ground truth solutions at time $T=50$ seconds, and models predictions for incompressible Navier-Stokes in vorticity form (high viscosity of $1e^{-3}$). \ourmethod{} reduces solution error w.r.t FNOs by over $20\%$ and FFNOs by over $40\%$. A single forward pass of \ourmethod{} models is on average $2 \times$ faster than FNO and $10\times$ than FFNOs. }
\label{fig:navier-stokes-predictions-big}
\end{figure*}

\paragraph{Hyperparameter tuning}
We start with the basic model structure of FNOs as detailed \citep{li2020fourier} and perform a basic hyperparameter search on a small slice of the training set, with the goal of ensuring proper convergence of a model. We did not find the number of layers to have a significant impact on convergence. Width plays an important role and is best kept above $24$.

\paragraph{Scaling laws}
We use the same settings as the main experiment, repeating separate training runs for the low viscosity setting. In particular, we increase the dataset size for each set of runs by a factor of $2$: $1024, 2048, 4096, 8192$. The total number of epochs is kept fixed, so that more iterations are performed for larger datasets. The same test set of size $200$ is used in all cases.

\paragraph{Further comments}
Additional predictions are provided in \cref{fig:navier-stokes-predictions-big}. \cref{fig:navier-stokes-modes-mae-psnr} shows the approximation error on the Navier-Stokes solutions due to truncation at different number of k-space elements $m$.

\subsection{Flow Around Airfoils}\label{asec:exp_dfp}
\paragraph{Dataset}
We use a slice of the dataset introduced by \cite{thuerey2020deep} in the form of $11000$ training pairs of initial conditions and solutions. The solutions are obtained via {\tt OpenFOAM} \citep{jasak2007openfoam} SIMPLE, a steady-state solver for incompressible and turbulent flows. In particular, the initial conditions are specified as freestream velocities over the domain (two-directional components), in addition to a specification of the airfoil in point cloud format. Delaunay triangulation is used for mesh generation.

After simulation, data is provided as initial condition and steady-state solution pairs. The initial condition is a three channel $128 \times 128$ image: two channels for freestream velocities and one for the airfoil mask. The solution is a three channel $128 \times 128$ image: a velocity field and a scalar pressure field. All data is normalized using training set statistics.

\paragraph{Models and training}
Training configuration is given as

\begin{listing}[H]
\begin{mintedbox}{yaml}
datamodule: 
    ntrain: 8000
    nval: 2000
    ntest: 1000
    batch_size: 64
train:
    optimizer: 
        type: AdamW
        learning_rate: 1e-3
        weight_decay: 1e-4
    scheduler: 
        type: Step
        step_size: 100
        gamma: 0.6
        scheduler_interval: epoch
loss_fn: RelativeL2Loss
\end{mintedbox}
\vspace{-6mm}
\end{listing}

The baseline UNet matches the architecture of \citep{thuerey2020deep} (DFPNet). The FNO architecture is comprised of a standard stack of FDM layer as discussed in \ref{asec:exp_nvs}. The k-space UNet in \ourmethod{+} has the same structure as a DFPNet.

\begin{listing}[H]
\begin{minipage}[t]{0.32\textwidth}
\begin{mintedbox}{yaml, title=\config{FNO}}
FNO:
modes: 24 
nlayers: 6
width: 48  \end{mintedbox}
\end{minipage}
\begin{minipage}[t]{0.32\textwidth}
\begin{mintedbox}{yaml, title=\config{DFPNET}}
DFPNET:
channel_exponent: 6 \end{mintedbox}
\end{minipage}
\begin{minipage}[t]{0.32\textwidth}
\begin{mintedbox}{yaml, title=\config{\ourmethod{+}}}
T1+: 
modes: 100 
channel_exponent: 5 \end{mintedbox}
\end{minipage}
\vspace{-6mm}
\end{listing}

\begin{figure*}[h!]
    \centering
    \includegraphics[width=\linewidth]{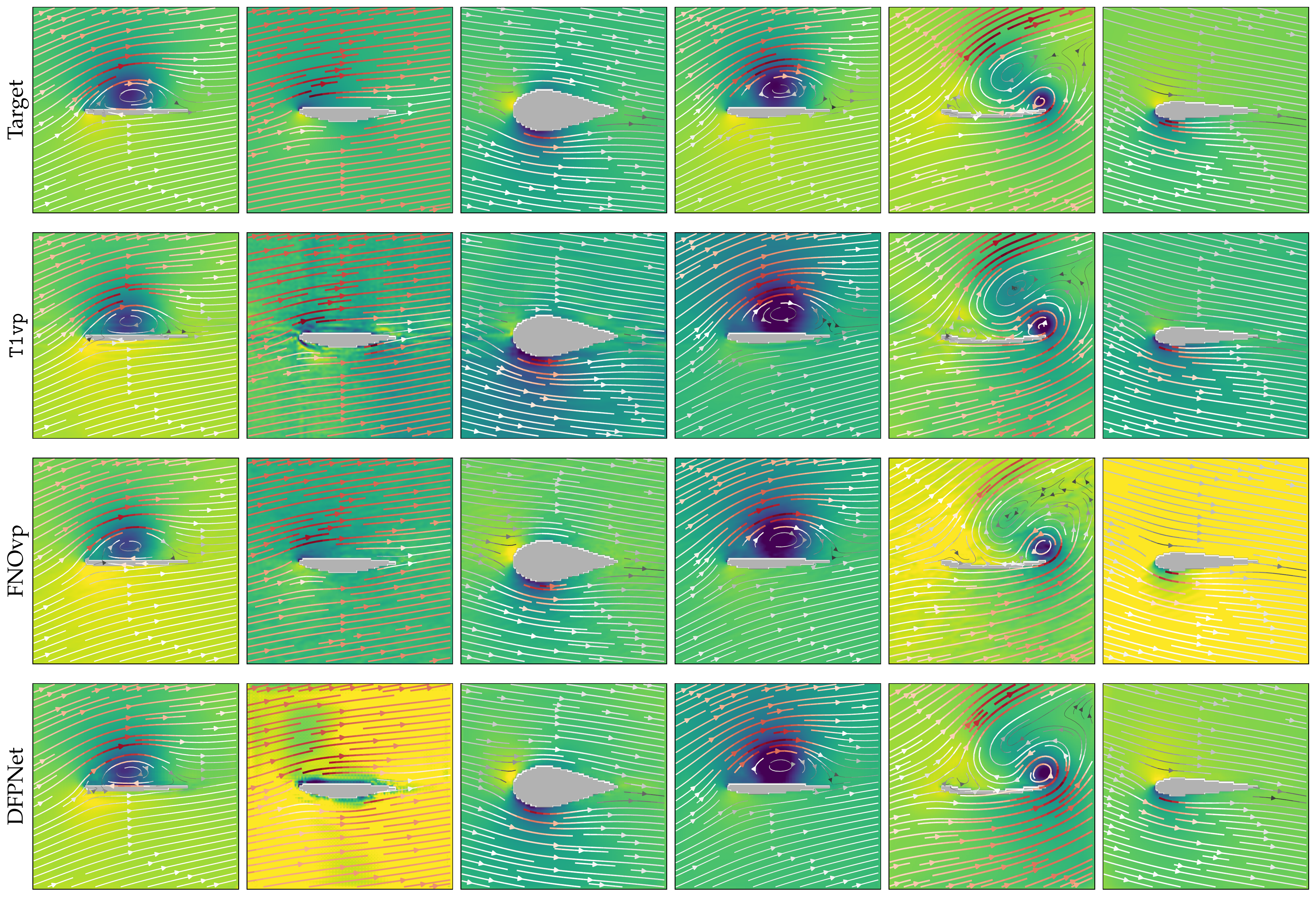}
    \vspace{-4mm}
    \captionof{figure}{\small Ground truth solutions and predictions with different airfoil designs and angles of attack of the flow. The background color is the scalar pressure value while the vector field represents the velocity field: arrow colors indicate its "strength" i.e. $2$-norm.}
\label{fig:dfp-predictions-big}
\end{figure*}

\begin{figure*}[h!]
    \centering
    % \vspace{-10mm}
    \includegraphics[width=0.75\linewidth]{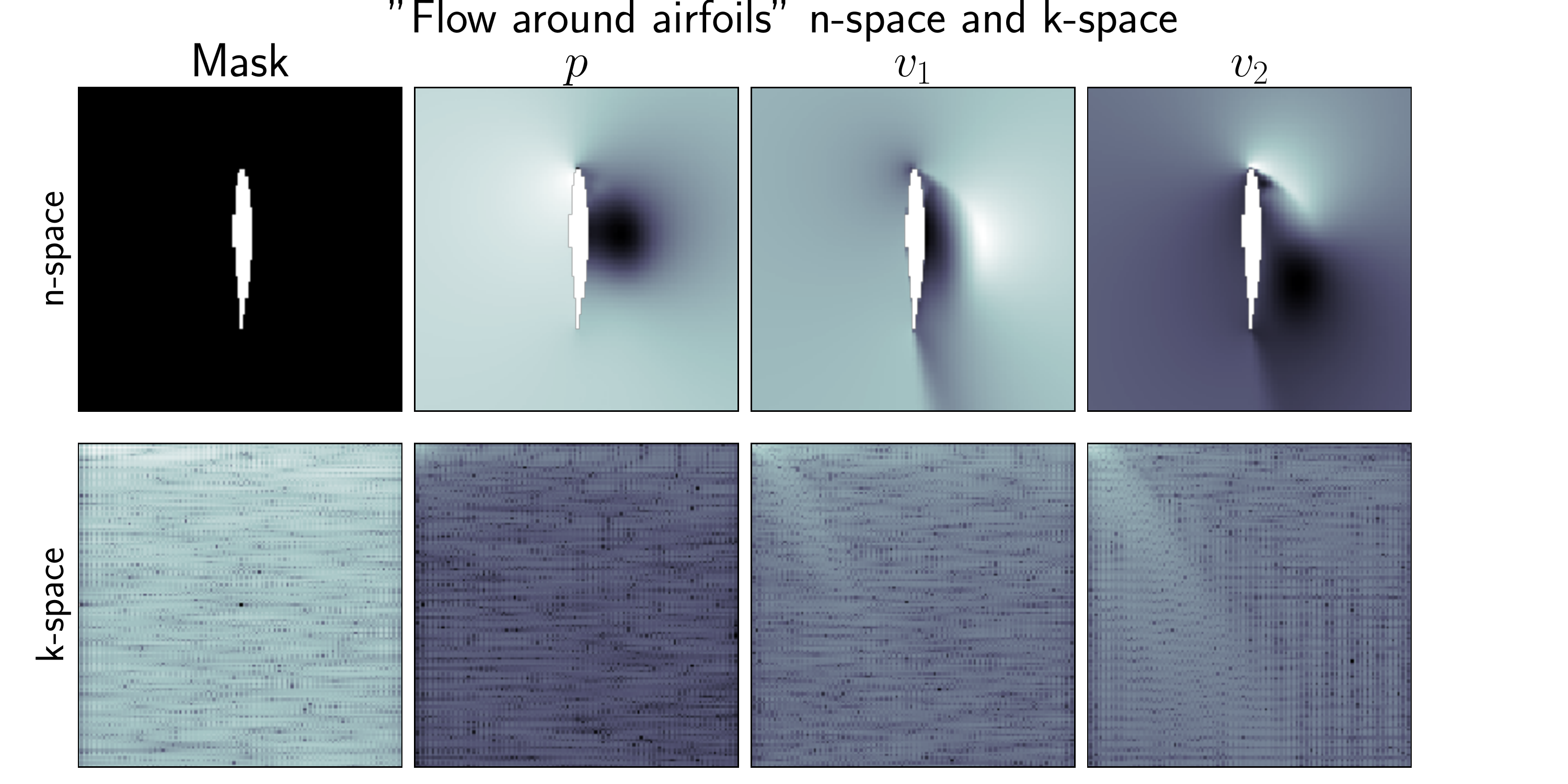}
    % \vspace{-3mm}
    \vspace{-3mm}
    \caption{\small Flow around airfoils: example of n-space: input mask, output pressure $p$ and velocity field $v_1, v_2$. Below, the corresponding DCT k-space in abs-log i.e. $\log{(|\cT(x)|)}$ to highlight its structure.}
    \label{fig:dfp-nspace-kspace}
\end{figure*}

\begin{figure*}[h!]
    \centering
    % \vspace{-10mm}
    \includegraphics[width=1\linewidth]{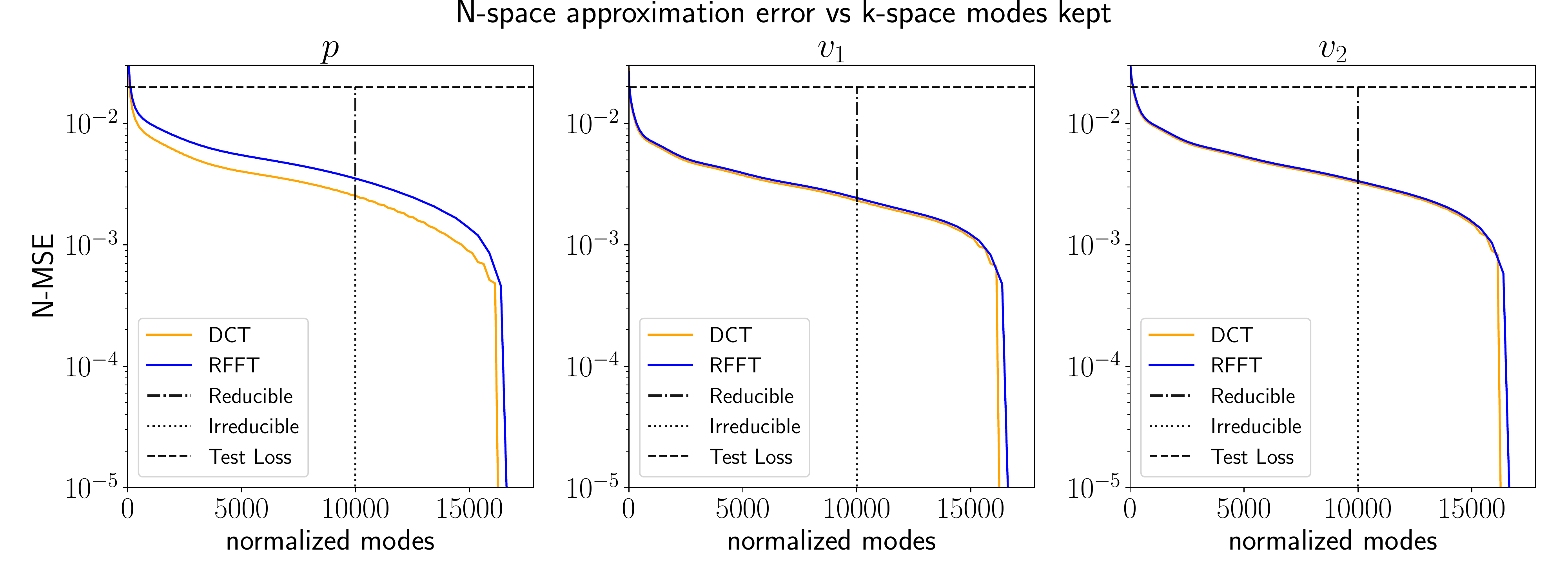}
    % \vspace{-3mm}
    \vspace{-8mm}
    \caption{\small Average approximation error (N-MSE) due to truncation in k-space at different number of elements $m$ for the \textit{flow around airfoils} dataset. In blue, the real FFT k-space, in orange the regular DCT k-space. On the x-axis, the normalized cost for a number of modes $m$: for DCTs, since the k-space is real, truncation at $m$ modes requires $m^2$ floats, for real FFTs with complex k-space and conjugacy the cost in floats is $4m^2$. The vertical line indicates the budget used for \ourmethod{} used in this task ($m=100$), while the horizontal line is the test N-MSE achieved.} 
    \label{fig:dfp-approx-decay}
    
\end{figure*}

\paragraph{Hyperparameter tuning}
This is an example of a dataset where the k-space is full due to discontinuity in the solution given by the airfoil mask.

We use the training and validation sets to inspect the k-space and set $m$ to $100$ for the irreducible loss term to be sufficiently small as shown in \cref{fig:dfp-approx-decay}. We swept over $m$ for FNOs and found larger than $24$ to perform worse, likely due to k-space convolution being sufficient to capture higher frequency components. We observe DFPNets with larger channel exponents perform worse due to overfitting. 

\paragraph{Further comments}

A sample of predictions is given in \cref{fig:dfp-predictions-big}. \cref{fig:dfp-nspace-kspace} shows the n-space and corresponding DCT k-space of a data point. As can be observed, the k-space is structured but full due to the discontinuity caused by the airfoil mask. \cref{fig:dfp-approx-decay} shows the approximation error on solution fields due to truncation in k-space at different $m$. In this task, the DCT is more efficient, given a budget of modes to keep, as it yields lower errors. This error provides a theoretical lower bound for the predictive error achievable by a \ourmethod{} model with a given budget, reachable only if the \ourmethod{} predicts the first $m$ modes perfectly.

The vertical line indicates the budget used for the main text \ourmethod{} experiments ($m=100$), and the horizontal one the test N-MSE achieved. Various segments of the vertical line indicate reducible and irreducible components of the loss as discussed in \cref{subsec:inverse}. The theoretical limit at $m=100$ is well below what has been empirically achieved by \ourmethod{} and other models. Indeed, the irreducible loss is an order of magnitude smaller than what the best model (including non-reduced-order variants) achieves on the task.

\subsection{Turbulent Smoke}\label{asec:exp_sf}
\paragraph{Dataset}

We employ for this experiment the ScalarFlow dataset introduced in \citep{eckert2019scalarflow} which is available online under the Creative Commons license CC-BY-NC-SA 4.0\footnote{ScalarFlow dataset download: \href{https://ge.in.tum.de/publications/2019-scalarflow-eckert/}{https://ge.in.tum.de/publications/2019-scalarflow-eckert/}}. \cite{eckert2019scalarflow} created an environment for controlling the release of smoke plumes: a fog machine generated fog inside of a container; the fog was then heated up by a heating cable and a valve controlled its release. Data was captured via multiple calibrated cameras in high resolution at $60$ fps (frames per second) for $150$ frames. 

\begin{figure*}[h!]
    \centering
    % \vspace{-10mm}
    \includegraphics[width=0.8\linewidth]{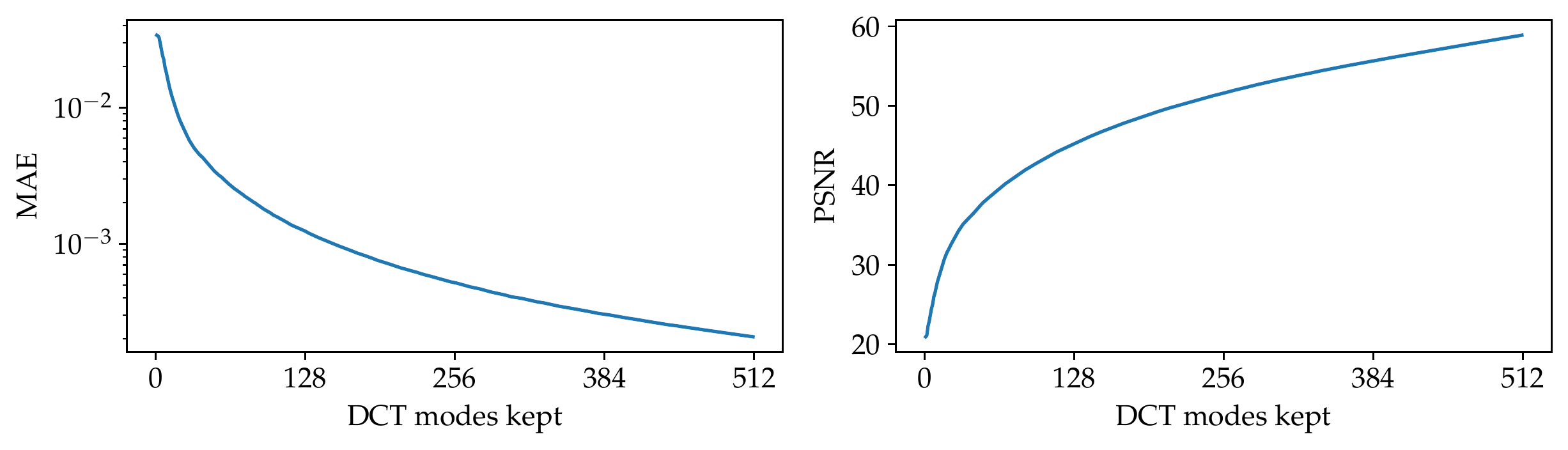}
    % \vspace{-3mm}
    \caption{\small ScalarFlow dataset: reconstruction error versus number of kept DCT modes.}
    \label{fig:scalarflow-modes-mae-psnr}
    \vspace{-3mm}
\end{figure*}

The dataset contains 3D reconstructions of the smoke plumes and 2D input and rendered images: input images are used by \cite{eckert2019scalarflow} to solve an optimization problem in which the goal is to generate a 3D reconstruction that minimizes the difference between input and rendered images. 2D input images are obtained directly from raw data on which only post-processing is applied by \citep{eckert2019scalarflow} in the form of gray scaling and denoising: these are saved in compressed $\tt numpy$ \citep{harris2020array} arrays named $\tt imgsTarget\_000xxx.npz$. Each resulting frame comprises $5$ different camera views $600\times1062$ in size. Since we want to use \ourmethod{} on high-resolution experimental data, we directly utilize the central camera view of these input images in our learning task without any further downsampling or data processing. Similarly to \citep{lienen2022learning}, we divide the  $104$ recordings into the first $64$ for training and use the remaining $20$ for validation and $20$ for testing. 

Data is normalized to the $[0,1]$ range based on training dataset statistics.

\begin{figure*}[h!]
    \centering
    \includegraphics[width=0.95\linewidth]{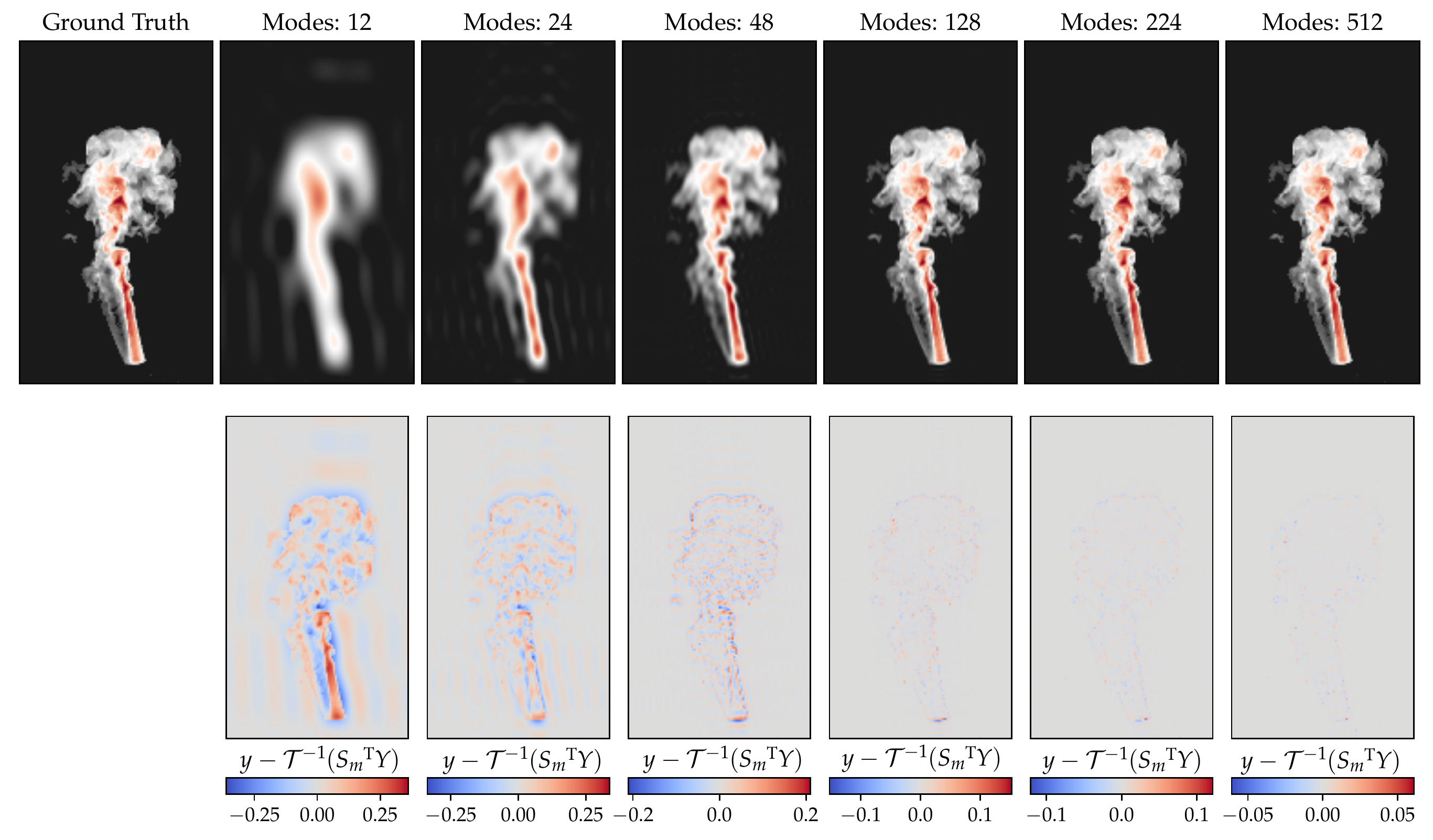}
    % \vspace{-3mm}
    \captionof{figure}{\small \textbf{[Top]} Visual comparison of ScalarFlow frames with changing number of DCT modes kept (i.e. first $m$ elements) .
    \textbf{[Bottom]} Error between the ground truth frame $y$ and its inverse transformation after mode pruning from $k$-space back to $n$-space. As expected, the first few k-space elements are crucial to minimizing reconstruction errors, with higher frequency components contributing minimally.}
\label{fig:scalarflow-modes-visualization}
\end{figure*}

\paragraph{Hyperparameter selection and tuning}
We performed a search on the most representative hyperparameters. One of the most important hyperparameters to choose from is the number of DCT modes to keep, i.e. first $m$ elements in $k$-space. We note that for simplicity as well as for compatibility with the UNet inside of \ourmethod{+}, we consider a  \textit{square} mode pruning, i.e. we keep the same number of frequencies on both height and width of the image and refer to the modes kept in both dimensions as $m$. \cref{fig:scalarflow-modes-visualization} and \cref{fig:scalarflow-modes-mae-psnr} show trends of DCT modes in terms of errors and visual quality: while the first modes $m$ contribute the most to the quality of the representation in $n$-space, the last elements contribute only to high-frequency details whose effect is minor on the overall reconstruction. Thus, we set \ourmethod{+} to $m=224$ and consequently \ourmethod{} to $m=512$ to have comparable model sizes. We set $m=48$ for FNO due to memory and model size limitations, noting that its residual connections effectively enlarge the training spectrum to all possible frequencies as shown in \cref{fig:scalarflow-comparison-raw}. Similarly to other experiments (B2), we observe raising $m$ in FNO to not significantly improve predictive error, even when the additional k-space elements would include a larger portion of the dataset.

We also experiment with different iterative rollout update strategies as in \citep{pfaff2020learning}. We consider the time step $\Delta t$ to be unitary, i.e. $\Delta t = 1$, given that the training frames are sampled consistently at 60 fps. We call $0$-order integration an update of the type: $x_{t+1} = h_\theta(x_t; x_{t-1}, \dots, x_{t-H})$ in which $h_\theta$ denotes a learned model which takes as inputs the current state $x_t$ and optionally a history of size $H$ of past states $x_{t-1}, \dots, x_{t-H}$ and directly predicts the next state $x_{t+1}$. A $1$-order integrator performs the following update: $x_{t+1} = x_t + h_\theta(x_t; \cdot)$, in which the model predicts the state update, i.e. the \textit{velocity}, similarly to an Euler step. A $2$-order integrator, also known as basic Störmer–-Verlet \citep{verlet1967computer} can be written as following: $x_{t+1} = 2x_t - x_{t-1} + h_\theta(x_t; \cdot)$; the model $h_\theta$ predicts the \textit{acceleration} of the system. We empirically found the zero-order integration to be more prone to generating artifacts with slower convergence, which may be because the model has to directly predict the next step with no "help" from the current step information. We found models trained with first-order integrators to have lower predictive errors than those trained with second-order ones, and we thus use it in all the experiments. As for the history size, we selected $H=1$ since it provided noticeable benefits compared to $H=0$, in which the model has no way of knowing previous states and thus inferring velocities. Larger history sizes did not seem to provide any improvements and only made the models larger as also noted in \citep{pfaff2020learning}.  

\begin{figure}
    \centering
    \includegraphics[width=\linewidth]{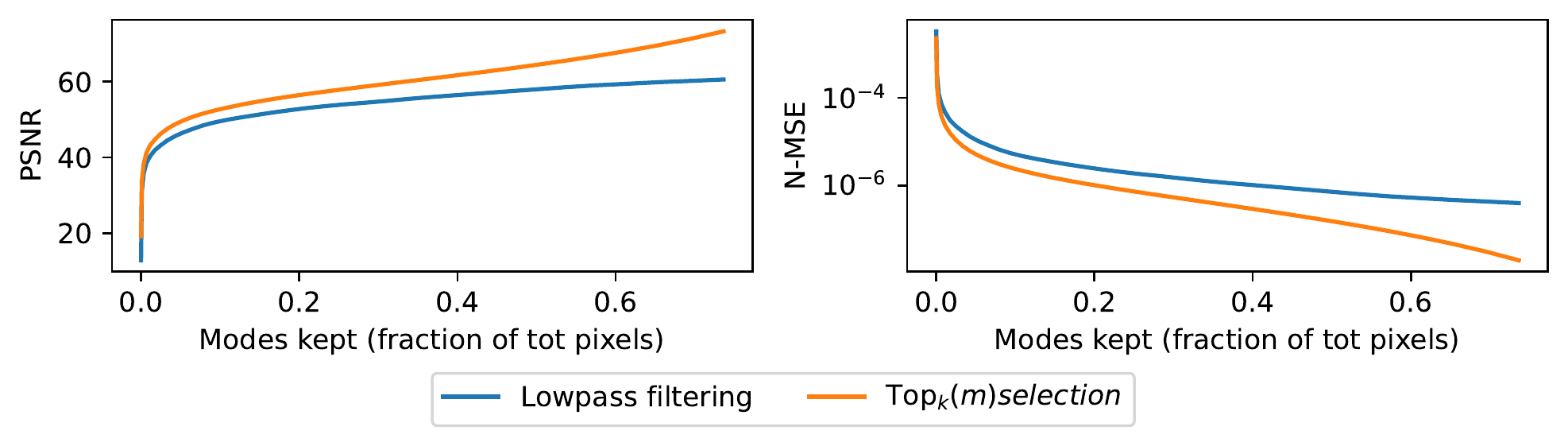}
    \vspace{-5mm}

    \caption{\footnotesize Reconstruction errors in pixel space of low-pass filtering of the lowest $m$ frequency modes vs ${\tt top}_k(m)$ selection on a single frame of ScalarFlow.}
    \vspace{-5mm}
    \label{fig:sflow_topk}
\end{figure}

\paragraph{Mode selection} We further show in \cref{fig:sflow_topk} the effect of simple low-pass filtering of lowest $m$ frequency modes and ${\tt top}_k(m)$ mode selection in pixel space reconstruction (as a fraction of total pixes, i.e., $600 \times 1062$). The latter achieves better reconstruction results with the same number of parameters.

% Trick for the configs
% If this environment is on the first element of a page, it breaks - make sure this doesnt happen
% \pagebreak % remove if not needed!!
\pagebreak
\paragraph{Models and training}

All models share the configuration for training:

\begin{listing}[H]
\begin{mintedbox}{yaml}
datamodule: 
    ntrain: 64
    nval: 20
    ntest: 20
    batch_size: 1
    history_size: 1
    target_steps_train: 3
    target_steps_val_test: 10
train:
    optimizer: 
        type: AdamW
        learning_rate: 1e-3
        weight_decay: 1e-4
    scheduler:
        type: CosineAnnealingWarmRestarts
        T_0: 32
        step_size: 1
        scheduler_interval: step
loss_fn: RelativeL2Loss
\end{mintedbox}
\vspace{-6mm}
\end{listing}

Where we used the implementation in $\tt PyTorch$ of the cosine annealing schedule with warm restarts\footnote{We used the scheduler \href{https://pytorch.org/docs/stable/generated/torch.optim.lr_scheduler.CosineAnnealingWarmRestarts.html}{$\tt torch.optim.lr\_scheduler.CosineAnnealingWarmRestarts$} with the number of iterations for the first restart $T\_0 = 32$. All other hyperparameters are the same as in the reference implementation.}.
The FNO architecture comprises a standard stack of FDM layers as discussed in B.1. The k–space UNet in \ourmethod{+} (and in its $\tt vp$ variant) has the same structure as a DFPNet.

\begin{listing}[H]
\begin{minipage}[t]{0.32\textwidth}
\begin{mintedbox}{yaml, title=\config{FNO}}
modes: 48
nlayers: 4
width: 48 \end{mintedbox}
\end{minipage}
\begin{minipage}[t]{0.32\textwidth}
\begin{mintedbox}{yaml, title=\config{\ourmethod{}}}
modes: 512
nlayers: 4
width: 8  \end{mintedbox}
\end{minipage}
\begin{minipage}[t]{0.32\textwidth}
\begin{mintedbox}{yaml, title=\config{\ourmethod{+}}}
modes: 224
nlayers: 1
width: 4
channel_exponent: 7 \end{mintedbox}
\end{minipage}
\vspace{-6mm}
\end{listing}

where we note that all models employ $\tt GeLU$ \citep{hendrycks2016gaussian} activation functions between inner layers.

\paragraph{Analysis of results}
\begin{figure*}[h!]
    \centering
    % \vspace{-10mm}
    \includegraphics[width=0.4\linewidth]{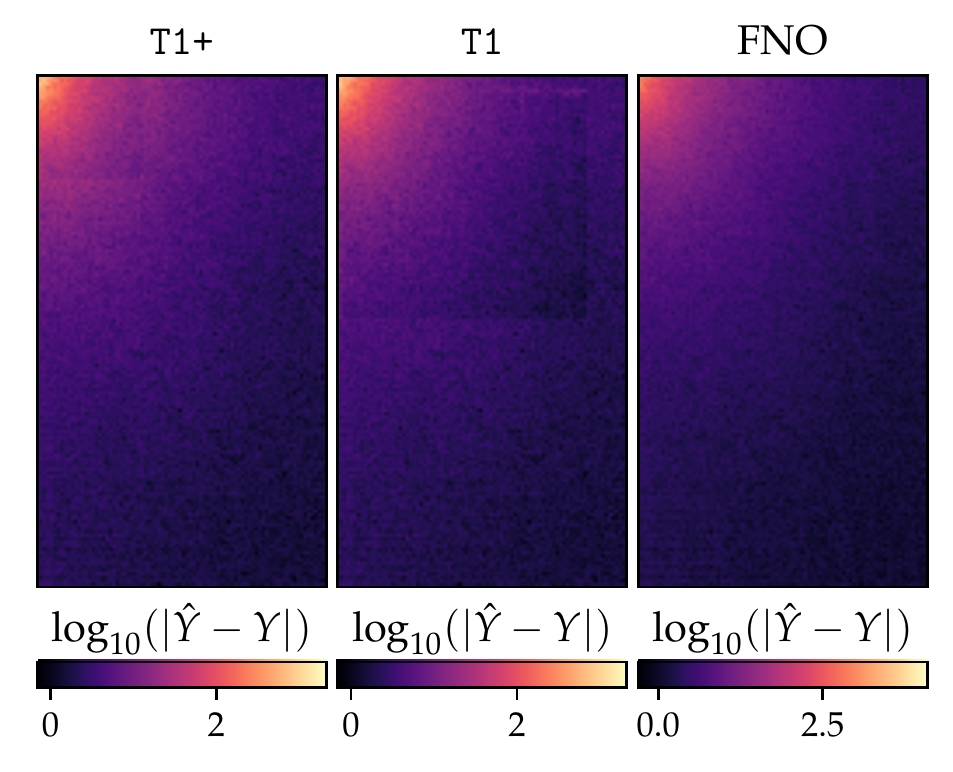}
    \vspace{-3.5mm}
    \caption{\small Mean log-absolute values of predictions in $k$-space (DCT-II) of a $20$-elements batch in the test dataset. Although \ourmethod{} is limited to $m=512$ and \ourmethod{+} to $m=224$ $k$-space elements (visible as square "shadows" in the error plots), its predictions are overall more physically accurate in $n$-space.}
    \label{fig:scalarflow-dct-batch-error}
\end{figure*}

\begin{table}[b]
    \caption{Full benchmark on the ScalarFlow dataset over 5 runs with different random seeds. N-MSE refers to 10-step test rollouts. \ourmethod{+vp} generates more stable rollouts while requiring a fraction of FNO's training time.}
    \centering
    \begin{tabular}{c|c c c c }\toprule
        \textbf{Method} & Param (M) & Size (MB) & Time (hrs) & N-MSE ($\times 10^{-1}$) \\
        \toprule
        FNO & 84.9 & 339 & 32.4 & 2.32 $\pm$ 0.02 \\ 
        \ourmethod{} & 83.9 & 335 & 8.1 & 2.39 $\pm$ 0.02\\ 
        \ourmethod{+} & 67.8 & 271 & 4.7 & 2.56 $\pm$ 0.16\\
        \ourmethod{+vp} & 67.8 & 271 & 4.7 & 2.28 $\pm$ 0.09\\
        \bottomrule
    \end{tabular}
    % \vspace{-1.4mm}
    \label{tab:scalarflow-large}
% \end{wraptable}
\end{table}

\cref{tab:scalarflow-large} provides a larger version of the table in the main text, including $1$-step mean absolute errors (MAE). We note that while FNO produces smaller errors in one-step predictions, it quickly accumulates larger errors in extrapolation. \cref{fig:scalarflow-dct-batch-error} shows mean errors in $k$-space of FNO vs \ourmethod{} and \ourmethod{+}. \ourmethod{} models demonstrate smaller overall errors and lower maxima compared to the FNO.

% \begin{wraptable}[0]{r}{\linewidth}

% Info
\section{Properties of Frequency Domain Models}
\subsection{Preliminary Results}
\begin{lemma}[Finite cosine series convergence] \label{lem:finite-cosine-series-simple}
    Let $k\in \mathbb{N}^+$, $N\in \mathbb{N}^+$ with $N \geq 2$. The following holds
    \begin{equation}
    \sum_{n=0}^{N-1} \cos(\frac{2 \pi k n}{N}) = 0.
    \end{equation}
\end{lemma}
\proof

Let us substitute $z = \frac{2 \pi k}{N}$ for simplicity. We can rewrite the finite series as follows

\begin{equation}
    y = \sum_{n=0}^{N-1} \cos(z n) = \cos(z \cdot 0) + \cos(z \cdot 1) + \dots + \cos(z (N-1)).
\end{equation}

By multiplying both sides of the equation by $2 \sin(z)$ we obtain

\begin{equation}
    \label{eq:cosine-series-expanded}
    2 \sin(z) y = 2 \cos(z \cdot 0) \sin(z) + 2 \cos(z \cdot  1) \sin(z) + \dots + 2 \cos(z (N-1)) \sin(z).
\end{equation}

By applying the following trigonometric identity

\begin{equation}
    2 \cos(\alpha) \sin(\beta) = \sin(\alpha + \beta) - \sin(\alpha - \beta),
\end{equation}

Equation \eqref{eq:cosine-series-expanded} becomes
\begin{equation}
    \begin{aligned}
    2 \sin(z) y =& ~2 \sin(z)\\
    &+ \sin(z + z) - \sin(z - z) \\
    &+ \sin(2z + z) - \sin(2z - z) \\
    &+ \sin(3z + z) - \sin(3z - z) \\
    &+ \dots \\
    &+ \sin((N-1)z + z) - \sin((N-1)z - z) \\
    \end{aligned}
\end{equation}

where terms on the right-hand side cancel out pairwise\footnote{Alternatively, we could think about the finite cosine series itself as the summation of $N$ cosine terms on a circle with terms from $0$ up to $N-1$ -- scaled by $k$, which does not affect the result. The cosine terms then cancel out in a pair--wise fashion (or in triplets, depending on even or odd $N$).}. After cleanup, we are left with the following 

\begin{equation}
    \begin{aligned}
        2 \sin(z) y =&  \sin(z) + \sin((N-1)z) + \sin(N z).
    \end{aligned}
\end{equation}

By substituting back $z = \frac{2 \pi k}{N}$ we obtain

\begin{equation}
    \begin{aligned}
        2 \sin(\frac{2 \pi k}{N}) \cdot y =&  \sin(\frac{2 \pi k}{N}) + \sin((N-1)\frac{2 \pi k}{N}) + \sin(N \frac{2 \pi k}{N})\\
        =&  \cancel{\sin(\frac{2 \pi k}{N})} - \cancel{\sin(\frac{2 \pi k}{N})} + \cancelto{0}{\sin(2 \pi k)},
    \end{aligned}
\end{equation}

where we used the trigonometric identity $\sin(-\alpha) = - \sin(\alpha)$. After dividing by the factor $ 2 \sin(\frac{2 \pi k}{N})$, we readily obtain the result $y = 0$.

\endproof
\begin{lemma}[Finite squared cosine series convergence] \label{lem:finite-cosine-series-squared}
    Let $k\in \mathbb{N}^+$, $N\in \mathbb{N}^+$ with $N \geq 2$. The following holds
    \begin{equation}
    \sum_{n=0}^{N-1} \cos^2 \left( \frac{2 \pi k n}{N} \right) = \frac{N}{2}.
    \end{equation}
\end{lemma}
\proof
We recall the following trigonometric identity

\begin{equation}
    \cos^2(\alpha) = \frac{1 + \cos(2\alpha)}{2}.
\end{equation}

Let us substitute $z = \frac{2 \pi k}{N}$ for simplicity. We can thus rewrite the finite series as follows

\begin{equation}
    \begin{aligned}
    \sum_{n=0}^{N-1}  \cos^2(z n )  &= \sum_{n=0}^{N-1} \frac{1 + \cos( 2 z n )}{2} \\
    &= \frac{1 + \cos(2 z \cdot 0)}{2} + \frac{1 + \cos(2 z \cdot 1)}{2} + \dots + \frac{1 + \cos(2 z (N-1))}{2} \\
    &= \frac{N}{2} + \frac{1}{2}  \left[ \cos(2 z \cdot 0) + \cos(2 z \cdot 1) + \dots + \cos(2 z (N-1)) \right] \\
    &= \frac{N}{2} + \cancelto{0}{\frac{1}{2} \sum_{n=0}^{N-1} \cos(2 z t)} \quad \text{(from Lemma \ref{lem:finite-cosine-series-simple})}\\
    &= \frac{N}{2}.
    \end{aligned}
\end{equation}
\endproof

\subsection{Statistics Under Fourier Transform}
There are various ways to show how probability measures and moments propagated under frequency domain transforms. We showcase two additional proof methods based on change of variables or explicit computation for simple input distributions.

\begin{lemma}[Central moment preservation under unitary linear operators]\label{pres}
    Let $x\sim p_x(x)$, $x\in\bC$ and let $\cT$ be a unitary linear operator. With $X = \cT(x)$, it holds
    \[
        p_X(X) = p_x(\cT^{-1}(X))
    \]
\end{lemma}
\proof
    The result follows immediately from the change of variables formula
    \[
        \begin{aligned}
            p_X(X) &= p_x(\cT^{-1}(X))\det \left[\frac{\dd}{\dd X}\cT^{-1}(X)\right]\\
            & = p_x(x),
        \end{aligned}
    \]
    being $\partial_X\cT(X)$ the Jacobian of $\cT$, since 
    $$\det\frac{\dd}{\dd X}\cT^{-1}(X) = \det\frac{\dd}{\dd X}\cT(X) = 1.$$
\endproof
%

%boh
%\begin{tcolorbox}[enhanced, colback=green!5, breakable, drop fuzzy shadow, frame hidden]
%
\begin{lemma}[Variance preservation under unitary linear operators]\label{explicit_vp}
Let $x\in\R^N$ be a random vector with 
\[
    \bE[x] = \0, \quad~ \mathbb{V}[x] = \sigma^2 \Id.
\]
with $\cT$ a normalized DFT. If $X = \cT(x)$, it holds
    \[
        \forall k,n: \quad \bE[X_k] = \bE[x_n] = 0 \quad \text{and} \quad \bV[X_k] = \bV[x_n] = \sigma^2
    \]
\end{lemma}
\proof

Let $x$ be real-valued input and distributed according to
\[
p_{\Re(x)} = \mathcal{N}(0, \sigma^2 \Id) \quad p_{\Im(x)} = \delta(\0).
\]

Consider a single element of $X$ 
\[
    X_k = \sum_{n=0}^{N-1} v_n
\]
with 
\[
    v_n = \frac{1}{\sqrt{N}}e^{\frac{2\pi jnk}{N}}x_n = \frac{1}{\sqrt{N}}\cos \frac{2\pi nk}{N}x_n + j\frac{1}{\sqrt{N}}\sin\frac{2\pi nk}{N}x_n.
\]
For clarity, we will treat the real part $\Re(X_k)$ first. 
\[
    \Re(v_n) = \frac{1}{\sqrt{N}}\cos \frac{2\pi nk}{N} \Re(x_n) 
\]
and 
\[
    \begin{aligned}
        \bE[v_n] &= \frac{1}{N} \cos^2{\frac{2\pi nk}{N}}\mathbb{E}[x_n] = 0\\
        \mathbb{V}[v_n] &= \frac{1}{N} \cos^2{\frac{2\pi nk}{N}}\mathbb{V}[x_n] = \frac{\sigma^2}{N} \cos^2{\frac{2\pi nk}{N}}  
    \end{aligned}
\]
where we have used the fact that 
\[
    \frac{1}{\sqrt{N}}\sin\frac{2\pi nk}{N}\Im(x_n) = 0.
\]
Thus,
\[
    \begin{aligned}
        \bE[\Re(X_k)] &= 0\\
        \mathbb{V}[\Re(X_k)] &= \sum_{n=0}^{N-1}\frac{\sigma^2}{N} \cos^2{\frac{2\pi nk}{N}}
    \end{aligned}
\]
We observe that (a) the first central moment is preserved and (b) the variance term can be simplified as 
\[
    \begin{aligned}
    \mathbb{V}[\Re(X_k)] &= \sum_{n=0}^{N-1}\frac{\sigma^2}{N} \cos^2{\frac{2\pi nk}{N}} \\
    &=\frac{\sigma^2}{N}\sum_{n=0}^{N-1} \cos^2{\frac{2\pi nk}{N}} \\
    &= \frac{\sigma^2}{N}\frac{N}{2}\quad \text{(from Lemma \ref{lem:finite-cosine-series-squared})} \\
    &= \frac{\sigma^2}{2}
    \end{aligned}
\]
We follow a similar procedure for $\Im(X_k)$, arriving at
\[
    \begin{aligned}
        \bE[\Im(X_k)] &= 0\\
        \mathbb{V}[\Im(X_k)] &= \sum_{n=0}^{N-1}\frac{\sigma^2}{N} \sin^2{\frac{2\pi nk}{N}}
    \end{aligned}
\]
where the variance again simplifies to
\[
    \sum_{n=0}^{N-1}\frac{\sigma^2}{N} \sin^2{\frac{2\pi nk}{N}} = \frac{\sigma^2}{2}
\]

Since $X_k = \Re(X_k) + j\Im(X_k)$,  
\[
    \begin{aligned}
        \bE[X_k] &= \bE[\Re(X_k)] + j\bE[\Im(X_k)] = 0 + j0 = 0\\
        \mathbb{V}[X_k] &= \bV[\Re(X_k)] + \bV[\Im(X_k)] = \sigma^2
    \end{aligned}
\]
%
% $$
% p_{X_k} = \mathcal{N}(0, \sigma^2) \implies p_{Z} = \mathcal{N}(0, \sigma^2 I_{n_x})
% $$
\endproof

A similar argument can be developed using basic properties of circular-symmetry of complex Normals.

% $$
% p_{\Im(X_k)} = \mathcal{N}(0, \sum_{t=0}^{T-1}\frac{\sigma^2 \sin^2{(\frac{2\pi k t}{T})}}{T})
% $$

It is critical that the normalization factor $\frac{1}{\sqrt{N}}$ be included in $W$ in order to preserve the variance of $\mathbb{V}[X]$.

Indeed, normalization factors used in different conventions lead to different results
$$
\begin{aligned}
&\textsf{forward factor}~~\frac{1}{N} \implies \mathbb{V}[X_k] = \frac{\sigma^2}{N} \\
&\textsf{backward factor}~~1 \implies \mathbb{V}[X_k] = N \sigma^2
    \end{aligned}
$$
As $N$ can easily be in the order of hundreds or thousands for generic signals, explosion of variance can be an issue if the orthogonalization factor $\frac{1}{\sqrt{N}}$ is not applied to $W$.
%

% \bibliographystyle{abbrvnat}
% \bibliography{bibliography/main.bib}

\end{document}